\documentclass{article}

\usepackage[final]{neurips_2023}

\usepackage{dblfloatfix}
\usepackage[utf8]{inputenc} 
\usepackage[T1]{fontenc}    
\usepackage{hyperref}       
\usepackage{url}            
\usepackage{booktabs}       
\usepackage{amsfonts}       
\usepackage{nicefrac}       
\usepackage{microtype}      
\usepackage{xcolor}    
\usepackage{hyperref}
\usepackage{wrapfig}

\usepackage{algorithm}
\usepackage{algpseudocode} 
\usepackage{multicol}
\usepackage{graphicx}
\usepackage{caption}
\usepackage{comment}
\usepackage{subcaption}
\usepackage{tikz}
\usepackage{xspace}
\usepackage{amsmath}

\newcommand{\fullup}{full-update\xspace} 
\newcommand{\modular}{modular-update\xspace} 
\newcommand{\head}{head-tuning\xspace} 
\newcommand{\Fullup}{Full-update\xspace} 
\newcommand{\Modular}{Modular-update\xspace} 
\newcommand{\Head}{Head-tuning\xspace} 
 
\newcommand{\local}{\texttt{Local-only}\text{ learning}\xspace} 
\newcommand{\FedAvg}{\texttt{FedAvg}\xspace} 
\newcommand{\yolo}{\texttt{FedYolo}\xspace} 
\newcommand{\Fedlocal}{\texttt{FedAvg+Local}\xspace} 
\newcommand{\numpara}{P}
\definecolor{mligreen}{RGB}{153, 204, 255}

\newcommand{\acc}{{\text{Acc}}}
\newcommand{\FL}{FL\xspace}
\definecolor{darkred}{RGB}{150,0,0}
\definecolor{darkgreen}{RGB}{0,150,0}
\definecolor{darkblue}{RGB}{0,0,200}
\newcommand{\darkred}[1]{\textcolor{darkred}{#1}}

\title{FedYolo: Augmenting Federated Learning with Pretrained Transformers}

\author{%
  Xuechen Zhang$^1$\quad Mingchen Li$^1$\quad Xiangyu Chang$^1$\quad
  Jiasi Chen$^{12}$
  \And
  Amit K.~Roy-Chowdhury$^1$ \quad
  Ananda Theertha Suresh$^3$\quad
  Samet Oymak$^{12}$
}

\begin{document}

\maketitle
\let\thefootnote\relax\footnote{$^{1}$ University of California, Riverside, $^{2}$ University of Michigan, Ann Arbor, $^{3}$ Google Research. Correspondence to: Xuechen Zhang <xzhan394@ucr.edu> and Samet Oymak <oymak@umich.edu>.}


\begin{abstract}
The explosive growth and diversity of machine learning applications motivate a fundamental rethinking of learning with mobile and edge devices. How can we address \emph{diverse/disparate client goals} and learn with \emph{scarce heterogeneous data}? While federated learning aims to address these issues, it has several bottlenecks and challenges hindering a unified solution. On the other hand, large transformer models have been shown to work across a variety of tasks often achieving remarkable few-shot adaptation. This raises the question: Can clients use a single general-purpose model -- rather than custom models for each task -- while obeying \emph{device and network constraints}?  In this work, we investigate pretrained transformers (PTF) to 
achieve these on-device learning goals and
thoroughly explore the roles of model size and modularity, where the latter refers to adaptation through modules such as prompts or adapters. 
Focusing on federated learning, we demonstrate that:
\textbf{(1) Larger scale} shrinks the accuracy gaps between alternative approaches and improves heterogeneity robustness.
Scale allows clients to run more local SGD epochs which can significantly reduce the number of communication rounds. At the extreme, clients can achieve respectable accuracy fully-locally highlighting the potential of fully-local learning.
\textbf{(2) Modularity}, by design, enables $>$100$\times$ less communication in bits. Surprisingly, it also boosts the generalization capability of local adaptation methods and the robustness of smaller PTFs. Finally, it enables clients to solve multiple unrelated tasks simultaneously using a single PTF, whereas full updates are prone to catastrophic forgetting. These insights on scale and modularity motivate a new federated learning approach we call ``You Only Load Once'' (FedYolo): The clients load a full PTF model once and all future updates are accomplished through communication-efficient modules with limited catastrophic-forgetting, where each task is assigned to its own module.
\end{abstract}

\section{Introduction}
\label{sect introduction}



Since its inception, federated learning (\FL) has enjoyed significant success as a versatile machine learning technique that enables collaboration across a large number of decentralized clients. On the other hand, the rich problem landscape of \FL poses multiple fundamental challenges \cite{li2020federated}. For instance, in typical \FL scenarios, limited client data and heterogeneity arising from variations in client distributions, create optimization bottlenecks. In parallel, rapid innovations in artificial intelligence applications prompts clients to use machine learning for a diverse set of tasks, even in mobile and edge computing. Heterogeneity \cite{gao2022survey,9134408}, and multitasking \cite{kumar2023mtfl,liu2023federated,he2022spreadgnn,ma2022state} are both active areas of research and are inherently linked to concerns such as catastrophic forgetting \cite{huang2022learn,shoham2019overcoming} (e.g.~when client updates override each other). Despite rich \FL literature, to the best of our knowledge, we still lack a clear unified strategy that overcomes these challenges. 

On the other hand, recent architectural innovations, in particular the advent of transformers, and the growing trend of pretraining these models on vast datasets advise a simple strategy: Equip clients with a large pretrained transformer (PTF), such as a foundation model \cite{bommasani2021opportunities}. Since large PTFs can be few-shot \emph{adapted} to a wide range of downstream tasks (i.e.~\textbf{power of scale} \cite{dosovitskiy2020vit,lester2021power}), they can provide warm-start for \FL and enable better adaptation to local client distributions. Crucially, while very large PTFs with billions of parameters cannot be deployed in mobile devices, innovations in mobile hardware (equipped with GPU/TPU) \cite{kovacs2021object} and advances in model compression/distillation \cite{huang2022compressing,tao2022compression,xu2022survey} will make it possible to deploy smaller, yet equally effective models on clients' devices. 

While this high-level strategy seems viable conceptually, can we actually realize such benefits in a multitask \FL setting with heterogeneous data and communication bottlenecks? For instance, communicating (billions of) parameters of a state-of-the-art language model at each \FL round is far from being feasible. In this work, together with \textbf{scale}, we identify the \textbf{power of modularity} to address these \FL-specific challenges. Modularity refers to the fact that PTFs can be adapted to a wide range of downstream tasks through parameter-efficient modules such as prompt or adapter tuning. The question is: In a federated setting, how should clients utilize these modules?
Modularity suggests an intuitive PTF-specific FL strategy that we call \modular rule: Clients keep their PTF frozen throughout the training, and only train and communicate modules (and a classifier head) to adapt their PTFs to a new task as depicted in Figure \ref{fig:yolo_fig}. This is in contrast to conventional \fullup rule where clients send and receive all parameters of the model.


To explore the full benefits of PTFs, we consider a flexible FL setting where clients have heterogeneous few-shot data that might stem from multiple distinct learning tasks (e.g.~text analysis vs image classification). We study three training schemes, each of which can be implemented in a modular- or full-update fashion: In \local, each client trains its model using its private data. \FedAvg is the standard aggregation scheme for federated learning. Finally, \Fedlocal is a popular personalization scheme where clients locally fine-tune their model following $\FedAvg$. The training strategies we explore are depicted in Figure \ref{fig:yolo_fig}. In a nutshell, our main message is: 

\hspace{7pt}\emph{Large PTFs with modular updates naturally enable communication-efficient, robust, multitask FL.}

This message generalizes well across different module choices (prompt, LoRA, adapter), pointing to the universal benefit of parameter-efficient FL. Specifically, we make the following contributions:

\begin{figure*}[t]
\centering
\includegraphics[scale=0.44]{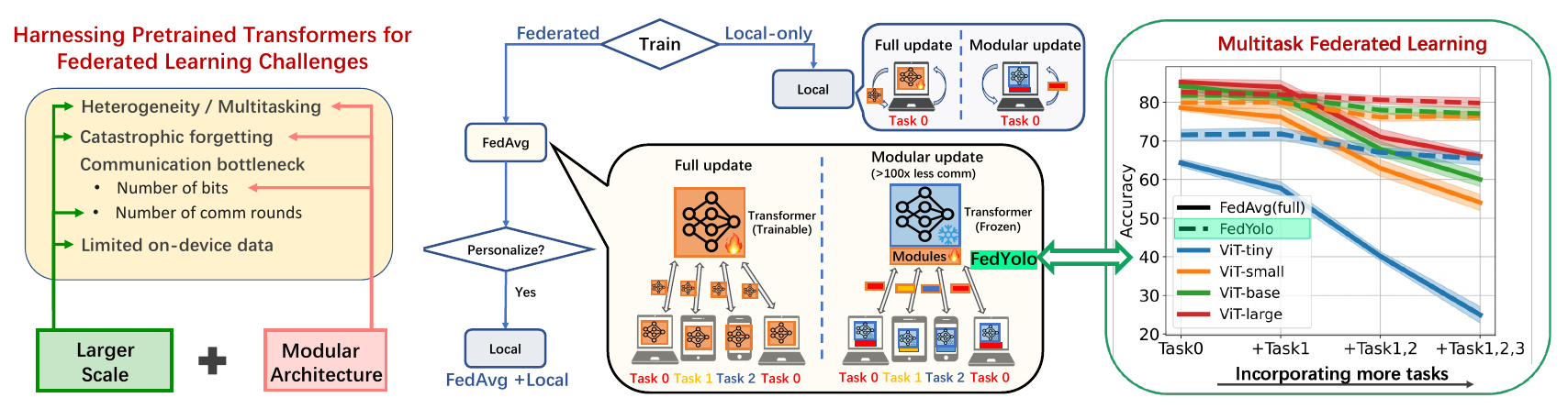}
\caption{\small{\textbf{Left side:} We investigate \emph{scale and modularity} of pretrained transformers (PTF) to address federated learning (FL) challenges.  \textbf{Center:} The training can branch into either FL or $\local$, once PTF model is loaded to the device. FL uses either $\FedAvg$ or $\Fedlocal$. All three training schemes could be implemented with two update methods: \Fullup and \modular as shown in the center box. {\Fullup trains and communicates all model parameters whereas \modular trains a small subset of parameters while freezing the PTF backbone.} 
\textbf{Right side:} For multitask FL, we propose $\yolo$ which assigns unique modules for each task (distinct colors of Tasks A,B,C). \yolo is superior to $\FedAvg$ (with full-update) as number of tasks grow thanks to modularity.}}
\label{fig:yolo_fig}\vspace{-14pt}
\end{figure*}
$\bullet$ \textbf{Need for collaboration / personalization.} Scale allows for better few-shot learning and reduces the reliance on personalization and collaboration by shrinking the accuracy gaps between \FedAvg and \Fedlocal as well as \Fedlocal and \local. We also found that modular-update often outperforms full-update under few-shot or heterogeneous data. This makes modular-update a surprisingly effective strategy for \local and \Fedlocal. Importantly, combined benefits of modularity and scale make \local fairly competitive with FL.



$\bullet$ \textbf{Heterogeneity.} We demonstrate and provide insights into how scale boosts robustness of \FL to data heterogeneity, while the modularity particularly improves the robustness of smaller PTFs. They also both provide resilience to forgetting: Accuracy of \Fedlocal remains competitive with \FedAvg on the global distribution even after the local-learning phase. 

$\bullet$ \textbf{Communication.} Number of module parameters are in the order of tens of thousands, thus, we find that, modular updates unlock orders-of-magnitude communication savings compared to full update. Crucially, this occurs while maintaining, and often accelerating, the rate of convergence in communication rounds.In synergy, larger scale significantly reduces the number of communication rounds by allowing clients to run more local SGD epochs without sacrificing global accuracy.


$\bullet$ \textbf{Multitask learning.} In a multitask setting where FL clients collaboratively and simultaneously learn multiple disparate tasks (e.g., classification on different domains such as CIFAR, CelebA, and FEMNIST datasets), the challenge is determining which parts of the model to update. Building on modularity and \emph{``one PTF for many tasks''}, we propose the \yolo algorithm (``You Only Load Once'') that assigns isolated modules to each task while keeping the PTF backend frozen. 
Fig.~\ref{fig:yolo_fig} (right side) demonstrates that \yolo performs on par with learning each task in isolation whereas multitasking with \fullup suffers from catastrophic forgetting even for large PTFs.


Our findings have important implications. Adapting large PTFs via modular-update not only provides a simple communication-efficient strategy with relatively minor drawbacks but also provides significant potential upsides in terms of personalization, robustness, and multitasking. Notably, scale and modularity makes \local a fairly competitive alternative to FL approaches \FedAvg or \Fedlocal, hinting at the viability of full privacy on the client side. Additionally, our proposal \yolo enables the clients to use a single PTF and multiple small modules to address diverse set of mobile ML goals, avoiding the need for maintaining/training multiple models.

\vspace{-5pt}\section{Related Work}\vspace{-5pt}
\label{sec:related}


\textbf{Federated Learning.} Data heterogeneity, multitasking, and personalization have been studied in \FL in various settings \cite{li2020federated,karimireddy2020scaffold,fallah2020personalized,lim2020federated,li2021fedrs}, in particular, heterogeneity has attracted significant attention in recent years. Much of the prior works focus on the design of algorithms rather than the model architecture. Closer to our work, \cite{tan2022federated} proposed FL with pretrained models and \cite{qu2022rethinking} identifies robustness benefits of transformer architecture over convolutional models, however these works only consider full updates whereas modularity is central to our message. Recent works \cite{zhang2022federated,zhao2022reduce} explore related modular-update ideas for federated learning for NLP problems, however don't explore the role of scale. As a major difference, they find that modular-update performs worse than full-update whereas we find that it can perform better in the few-shot or heterogeneous settings (which is often the case in practice). \cite{liu2022federated} focuses on computational efficiency of pretrained transformers via early-exit predictions. Importantly, ours is the only work that explores and provides a concrete solution for multitask learning with PTFs.





\textbf{Parameter-efficient tuning and pretrained transformers.} PTFs have garnered significant attention in machine learning, owing to their impressive performance across a wide variety of applications, including but not limited to computer vision and NLP \cite{brown2020language,dosovitskiy2020image}. Although non-federated, recent works explored the benefits of scale (model size as well as data and computation during pretraining) in robustness to forgetting \cite{ramasesh2022effect} and (few-shot) accuracy \cite{henighan2020scaling,zhai2022scaling,hernandez2021scaling}. 
Parameter-efficient tuning methods have shown significant promise for enabling lightweight adaptation of transformer architectures. By making adjustments to a limited set of parameters, these techniques avoid (potentially costly) modifications to the much larger backend architecture. There are two primary methods for PE tuning: (i) Training a subset of model parameters, usually done by placing a linear probe on top of pretrained features \cite{radford2021learning}, and (ii) integrating small modules within the network \cite{li2021prefix,ruckle2020adapterdrop,houlsby2019parameter,jia2022visual,hu2021lora}. For transformers, it is known that latter methods tend to achieve superior performance compared to head-tuning -- which we also demonstrate in the FL context (see supplementary). \cite{lester2021power} finds that the performance of prompt-tuning matches full-finetuning as the scale grows however it falls short for smaller transformers when tuning with full data. Complementing this, we find that modular methods can help small models as well under few-shot learning or heterogeneous data.





\vspace{-5pt}\section{Preliminaries and Experimental Setup}\label{sec:expsetup}



The objective of the next sections is to shed light on the fundamental benefits (as well as drawbacks) of the scale and modularity of PTFs.
These findings inform our \yolo algorithm in Section \ref{sec:exp_multitask}. We ask:\vspace{-10pt}
\begin{itemize}
\setlength\itemsep{1pt}
    \item How does the scale of PTF impact personalization and need for collaboration?(Section~\ref{sec:homogeneous})
    \item Do scale and modularity help with robustness to heterogeneous clients? (Section~\ref{sec:heterogeneous})
    \item How much can modular updates or scale improve communication costs over \fullup? Can they also help reduce number of communication rounds?(Section~\ref{sec:exp_efficiency})
    \item When there are multiple tasks, how should clients select the appropriate module for a task, on top of a common base PTF?
    (Section~\ref{sec:exp_multitask})
\end{itemize}
We provide the experimental setup details below:

\textbf{Datasets.} We conduct experiments on the CIFAR-10 and CIFAR-100 datasets \cite{krizhevsky2009learning}, as well as two real-world datasets (CelebA and FEMNIST) from the LEAF benchmark \cite{caldas2018leaf}, following \cite{qu2022rethinking, ramasesh2022effect}. Importantly, all our experiments focus on the few-shot setting where we train on subsets of these datasets. For instance, our CelebA and FEMNIST experiments use 2.6\% and 1.8\% fraction of the total sample size respectively.

$\bullet$ \emph{CIFAR-10 and CIFAR-100:} For federated learning, we have 20 clients inspired from the settings of \cite{qu2022rethinking,ramasesh2022effect}. To explore the performance under a limited sample size, we utilize a subset of the original training dataset. 
Experiments are conducted in both homogeneous and heterogeneous settings, where in the homogeneous setting, each client contains samples from all classes, and in the heterogeneous setting, each client contains samples from a subset of classes. We simulate three data partition sets and control the non-IID level by changing the number of classes included in each client. For the CIFAR-100 dataset, the ``mild heterogeneous'' data partition denotes 20 classes per client, while the ``more heterogeneous'' data partition denotes 5 classes per client. To ensure fair comparison across data partitions and meet the challenge of limited local data, we assign 100 samples to each client, regardless of the degree of heterogeneity. The data distribution of each local test set matches that of the local train set for each client. 
Further details are in the supplementary material~\ref{sec:partition}.


$\bullet$ \emph{CelebA and FEMNIST:} For CelebA, we partition the dataset onto the clients based on the celebrity in each photo and test on the binary classification task of smile presence. For FEMNIST, we partition the data based on the writer of the digit/character.
In accordance with \cite{caldas2018leaf,qu2022rethinking}, we increase the task difficulty by dropping clients with large number of samples (specifically, 8 samples for CelebA and 120 samples for FEMNIST). For each client, we partition the data into equal 50/50 train/test sets, so the class distribution of each local test set matches that of the local train set for each client.

\textbf{Pre-trained Transformer (PTFs):} In this study, all methods except for \fullup, employed frozen PTF backbones. We utilized four different scales of the Vision Transformer (ViT) architecture: ViT-large (ViT-L), ViT-base (ViT-B), ViT-small (ViT-S), and ViT-tiny (ViT-T). The models are pre-trained on ImageNet-21K from the official Google JAX implementation \cite{dosovitskiy2020vit, steiner2021augreg, rw2019timm}. To adapt to the number of classes for each dataset, a dataset-specific header is deployed. The number of trainable parameters for each training strategy is provided in the supplementary material~\ref{sec:module}: For ViT-Large, the full model has 300M parameters whereas module sizes vary from 300k to 900k.

\textbf{Modules:} We evaluated several modules for the \modular method, including Adapter~\cite{houlsby2019parameterefficient}, LoRA~\cite{hu2021lora}, and VPT~\cite{jia2022visual}. 
Due to space limitations, we only include the results of the Adapter method in the main paper, while the results of the LoRA and VPT methods are similar and relegated to the supplementary material~\ref{app:other_module}. Therefore, in the results below, the \modular and ``Adapter'' terms are used interchangeably.
To ensure a fair comparison, we deploy the modules on all transformer blocks, maintaining a fixed embedding dimension of 8 across different scales. The Appendix provides further details on the size of each module plus PTF. 

\textbf{Personalized training:} For heterogeneous data distribution (\S\ref{sec:heterogeneous}), we also perform personalized training after the global federated training. 
Each client will thus have its own personalized model.
During the personalized training, we fine-tune the average global model using local data to obtain a customized model for each client. 

\textbf{Evaluation metrics:} Unless otherwise stated, the evaluation of all models is based on the average local accuracy across clients. In the case of $\FedAvg$, the performance of the average global model is calculated and shared among all clients. For $\local$ and $\Fedlocal$, each client has its own fine-tuned model, so we compute the average performance of the individual models. In all figures, error bars correspond to one standard deviation. 

\textbf{Optimizers:} We use FedAvg with SGD optimizer, momentum parameter of 0.9, and no weight decay. The local training batch size is set to 32. In appendix, we also provide experiments for FedProx \cite{li2020federated} and FedDyn \cite{acar2021federated} which led to consistent conclusions as FedAvg (see Section \ref{sec:app_exp_deatil}).




\section{Experiments}
\label{sec:results}

\subsection{PTF Scale Boosts Performance}\label{sec:homogeneous} 
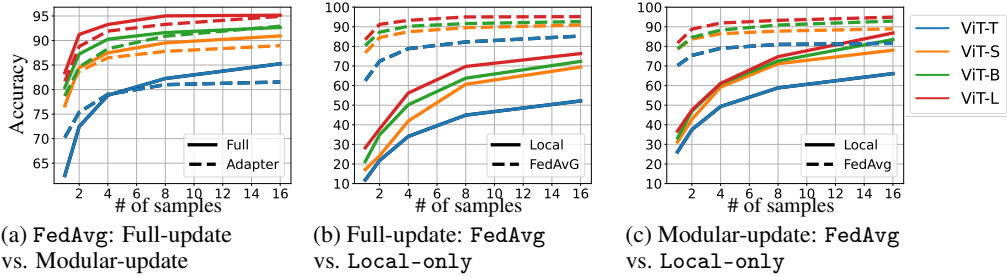
\begin{figure*}[t]
\centering
\vspace{-10pt}

\begin{subfigure}{1.6in}
    \centering
	\begin{tikzpicture}
		\node at (0,0) [scale=0.25]{\includegraphics{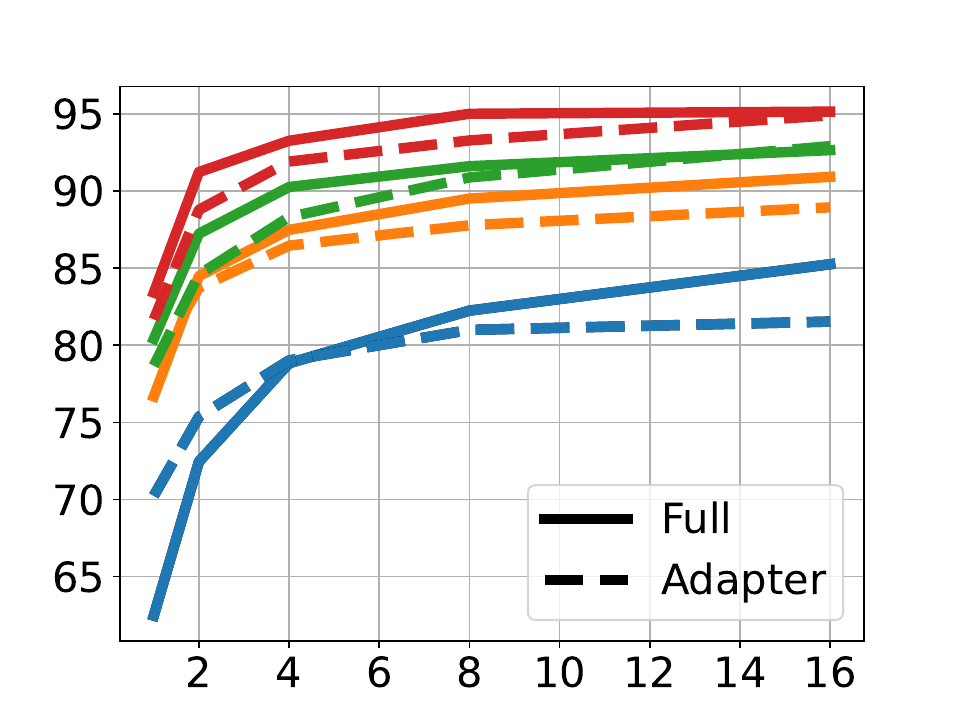}};
		\node at (-2 ,0) [scale=0.8,rotate=90] {Accuracy};
		\node at (0,-1.5) [scale=0.8] {\# of samples};
	\end{tikzpicture}\vspace{-6pt}\caption{\small{$\FedAvg$: \Fullup \\vs. \Modular}}\label{fig:cifar100_fed_fullmodule}
\end{subfigure}
\begin{subfigure}{1.6in}
    \centering
	\begin{tikzpicture}
		\node at (0,0) [scale=0.25]{\includegraphics{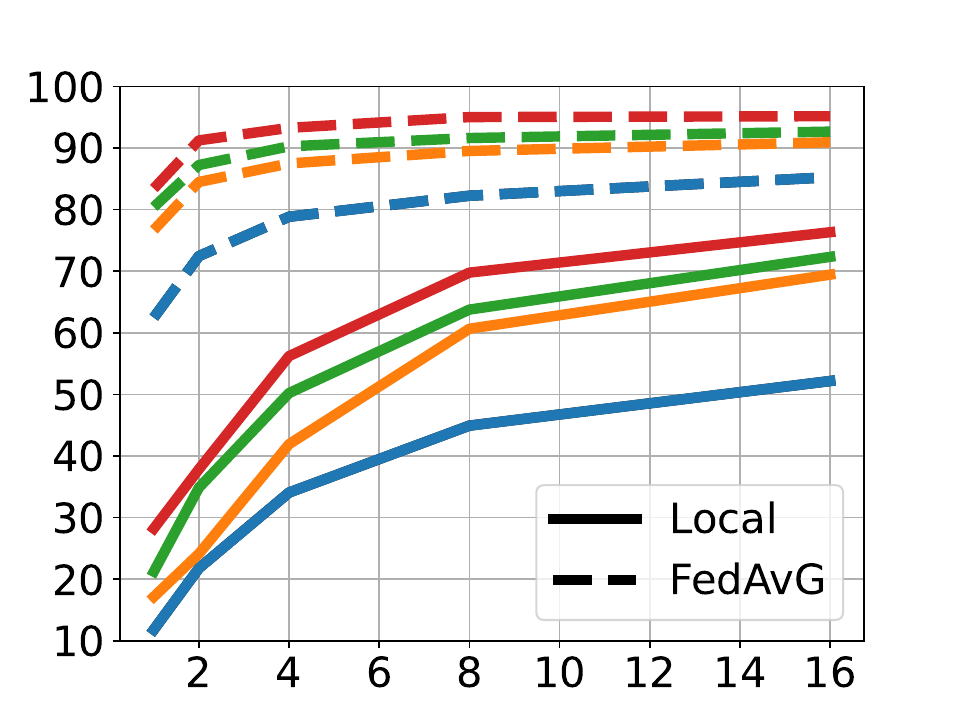}};
		\node at (0,-1.5) [scale=0.8] {\# of samples};
	\end{tikzpicture}\vspace{-6pt}\caption{\small{\Fullup: $\FedAvg$ \\vs. $\texttt{Local-only}$}}\label{fig:cifar100_full_fedlocal}
\end{subfigure}
\begin{subfigure}{1.6in}
    \centering
	\begin{tikzpicture}
		\node at (0,0) [scale=0.25]{\includegraphics{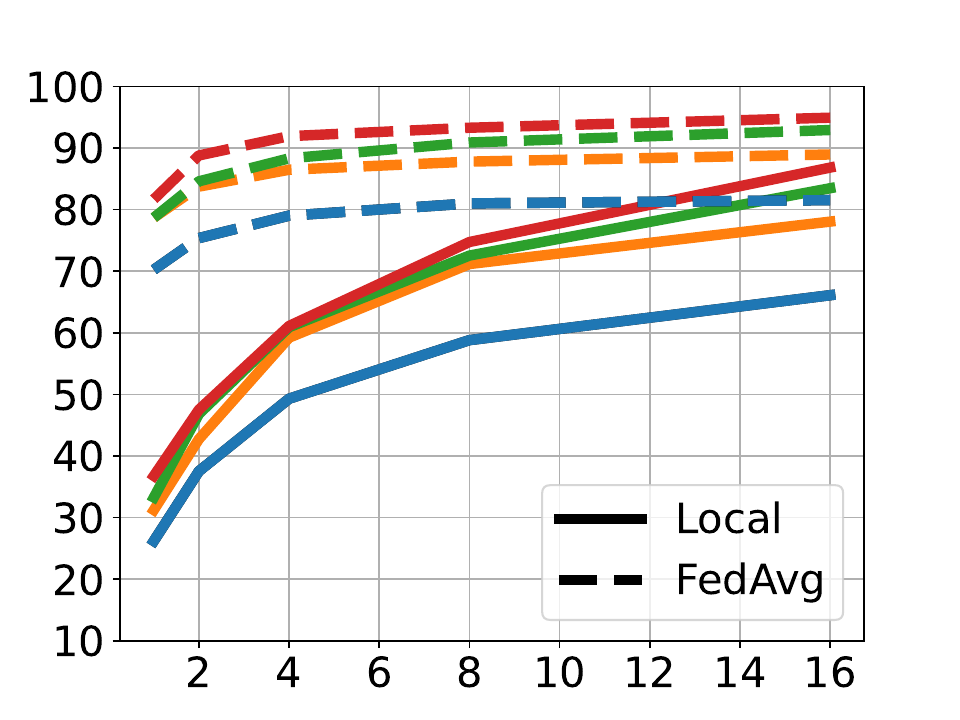}};
		\node at (0,-1.5) [scale=0.8] {\# of samples};
\end{tikzpicture}\vspace{-6pt}\caption{\small{\Modular: $\FedAvg$ \\vs. $\texttt{Local-only}$  }}\label{fig:cifar100_adapter_fedlocal}
\end{subfigure}
\hspace{-15pt}
\begin{subfigure}{0.15in}
    \centering
	\begin{tikzpicture}
		\node at (0,0) [scale=0.15]{\includegraphics{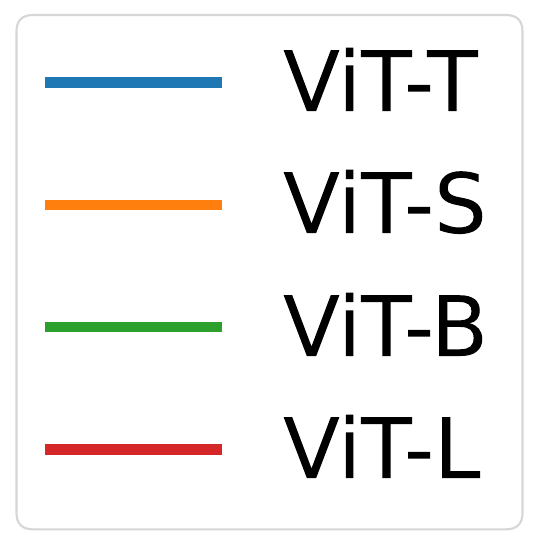}};
\end{tikzpicture}\vspace{60pt}
\end{subfigure}

\vspace{-5pt}
\caption{\small{Accuracy as a function of the number of training samples per class (CIFAR-100, all clients with 100 classes). 
(a): Larger PTFs improve accuracy for both \modular (dashed) and \fullup (solid) training strategies in the federated setting.
(b,c): Comparing a federated setting (\FedAvg, dashed) with a purely local setting (\local, solid), larger PTFs reduce the performance gap, especially with the \modular training strategy. 
}} 
\label{fig:fully_local}
\vspace{-10pt}
\end{figure*}

\begin{figure*}[t]
\centering
\begin{subfigure}{1.7in}
    \centering
	\begin{tikzpicture}
		\node at (0,0) [scale=0.29]{\includegraphics{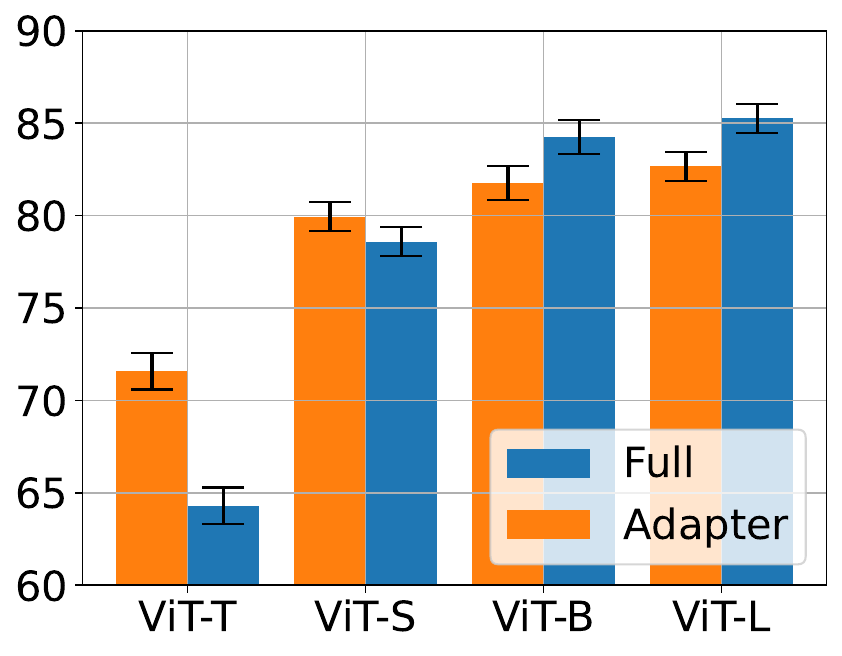}};
		\node at (-2.2 ,0) [scale=0.8,rotate=90] {Accuracy};
\end{tikzpicture}\vspace{-6pt}\caption{\small{CIFAR-100}}\label{fig:heter_cifar100}
\end{subfigure}
\begin{subfigure}{1.7in}
    \centering
	\begin{tikzpicture}
		\node at (0,0) [scale=0.29]{\includegraphics{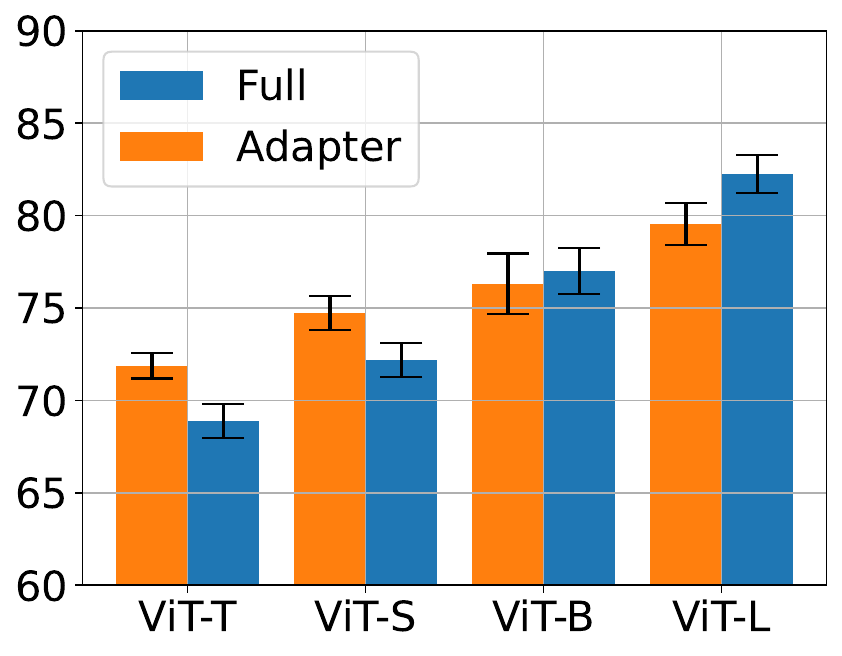}};
\end{tikzpicture}\vspace{-6pt}\caption{\small{CelebA}}\label{fig:heter_celeba}
\end{subfigure}
\begin{subfigure}{1.7in}
    \centering
	\begin{tikzpicture}
		\node at (0,0) [scale=0.29]{\includegraphics{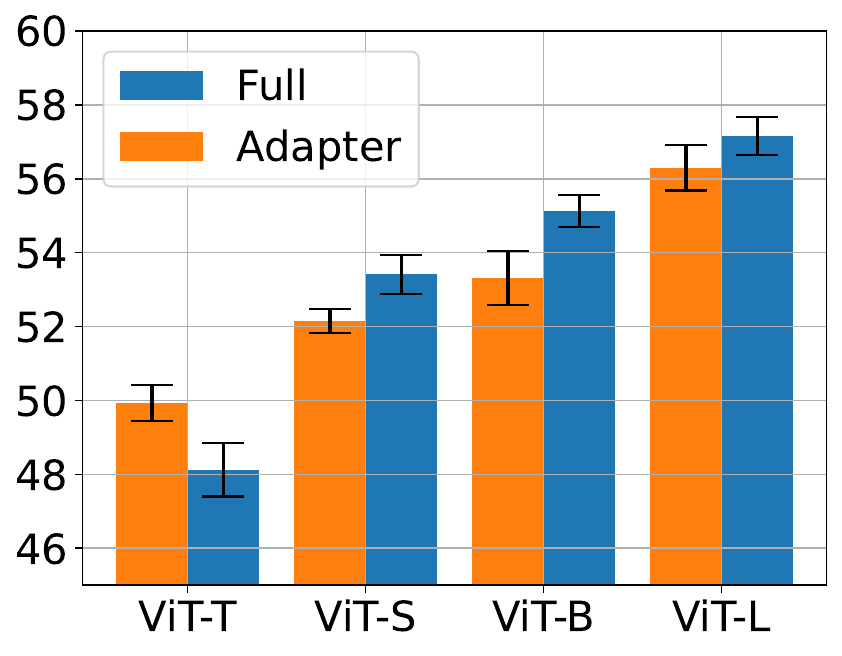}};
\end{tikzpicture}\vspace{-6pt}\caption{\small{FEMNIST}}\label{fig:heter_femnist}
\end{subfigure}
\vspace{-5pt}
\caption{\small{Comparison of different models with $\FedAvg$ aggregation and heterogeneous data distribution for different datasets. Larger PTFs (right side) outperform smaller PTFs (left side) whereas Adapter (\modular) are more helpful for smaller models and can outperform \fullup. }
}

\vspace{-10pt}
\label{fig:heter_acc}
\end{figure*}

\begin{figure*}[t]
\centering
\begin{subfigure}{1.2in}
    \centering
	\begin{tikzpicture}
		\node at (0,0) [scale=0.2]{\includegraphics{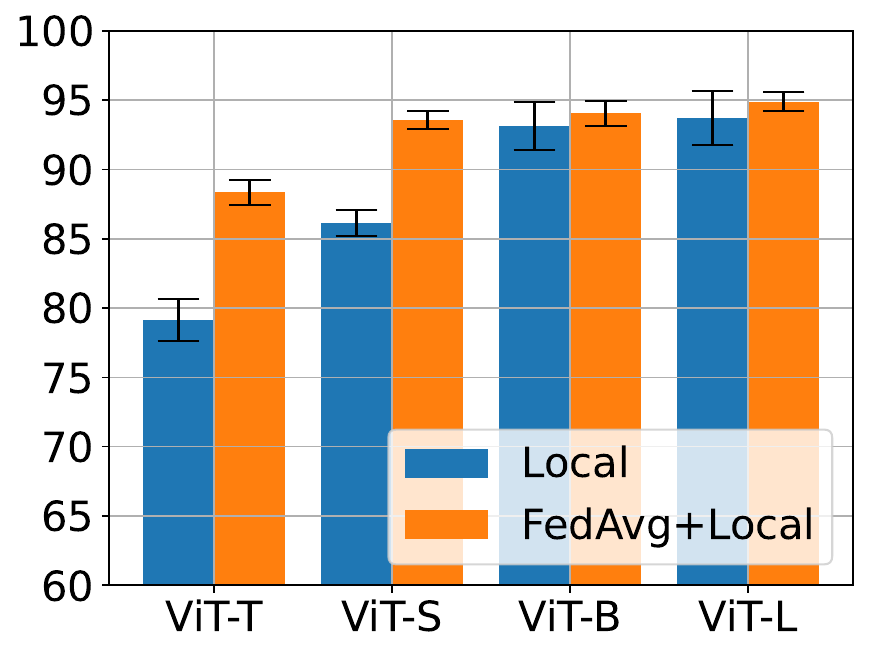}};
		\node at (-1.5 ,0) [scale=0.8,rotate=90] {Accuracy};
\end{tikzpicture}\vspace{-6pt}\caption{\small{CIFAR-100,\\\modular}}\label{fig:cifar100_adapter_localpersonalize}
\end{subfigure}
\begin{subfigure}{1.2in}
    \centering
	\begin{tikzpicture}
		\node at (0,0) [scale=0.2]{\includegraphics{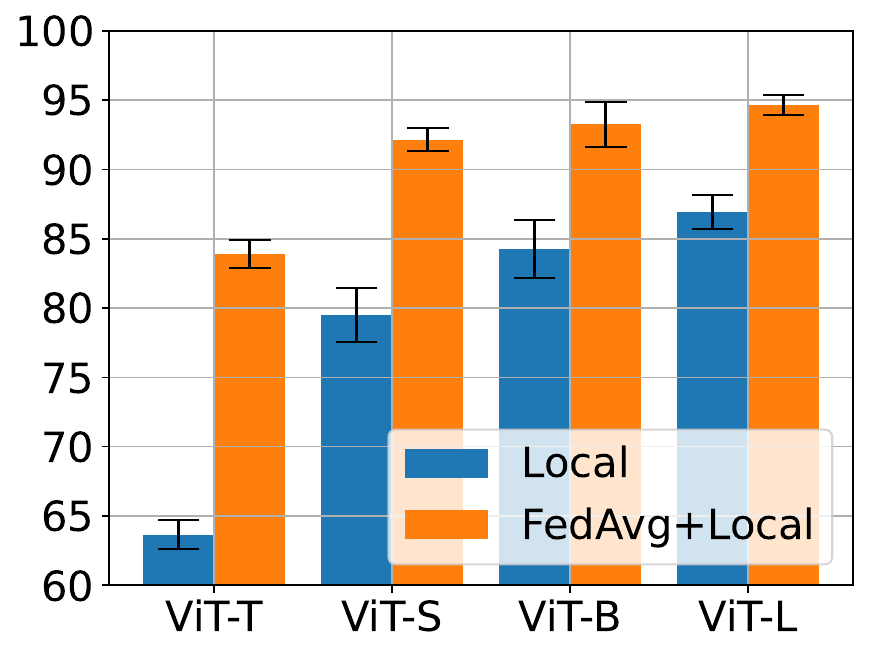}};
\end{tikzpicture}\vspace{-6pt}\caption{\small{CIFAR-100,\\\fullup}}\label{fig:cifar100_fully_localpersonalize}
\end{subfigure}
\begin{subfigure}{1.2in}
    \centering
	\begin{tikzpicture}
		\node at (0,0) [scale=0.2]{\includegraphics{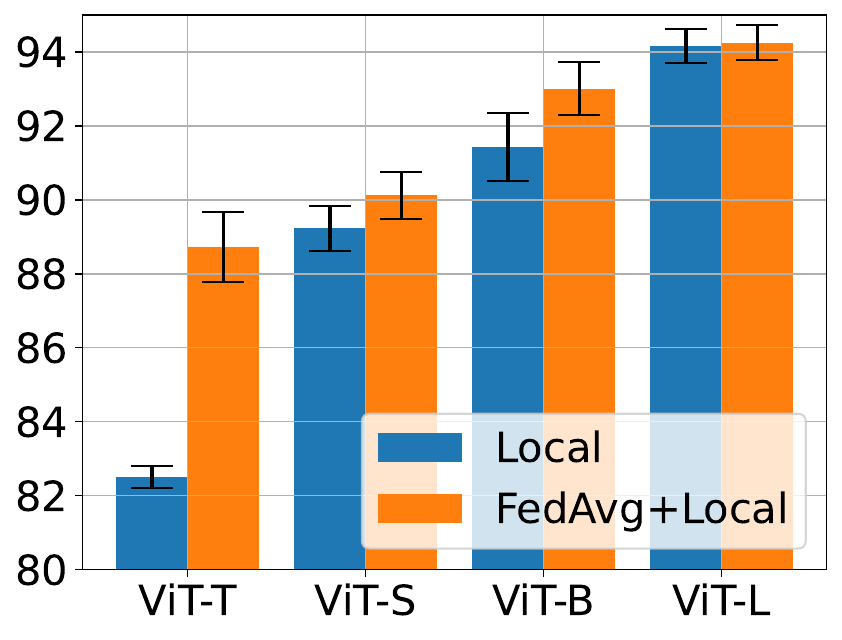}};
\end{tikzpicture}\vspace{-6pt}\caption{\small{CelebA, \\\modular}}\label{fig:celeba_adapter_localpersonalize}
\end{subfigure}
\begin{subfigure}{1.2in}
    \centering
	\begin{tikzpicture}
		\node at (0,0) [scale=0.2]{\includegraphics{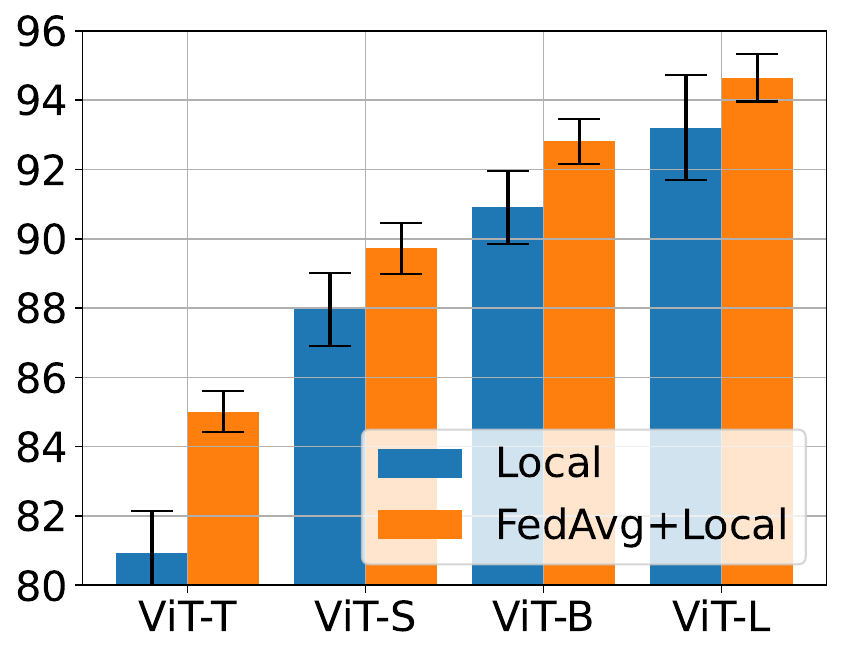}};
\end{tikzpicture}\vspace{-6pt}\caption{\small{CelebA,\\ \fullup}}\label{fig:celeba_fully_localpersonalize}
\end{subfigure}
\vspace{-5pt}
\caption{\small{Performance comparison between $\local$ (blue) and $\Fedlocal$ (orange), for different datsets
and update strategies (\fullup and \modular). As PTF scale increases, local training become more comparable to the federated setting.
}}
\label{fig:fulllocal_heter}
\vspace{-5pt}
\end{figure*} 
\textbf{Larger PTFs improve model performance:} 
In a federated setting, the presence of heterogeneous data distributions and limited samples among clients can lead to challenges in achieving optimal performance. 
The generalization benefits of large models (a.k.a.~over-parameterization) has been explored empirically as well as theoretically \cite{zhang2021understanding}. 
However, the impact of scale in FL is an underexplored topic and it is not immediately clear whether large models will retain their benefits in FL setting with local training and limited samples.
To study this, we conduct experiments in a federated few-shot setting, exploring both homogeneous and heterogeneous data distributions to understand the effects on model performance. For a homogeneous data distribution where all clients have the complete set of 100 classes from CIFAR-100, we plot the accuracy as the number of samples per class increases in Fig.~\ref{fig:cifar100_fed_fullmodule}.
We can see that larger PTFs consistently outperform smaller PTFs, regardless of whether the \fullup or \modular method is employed. 

\begin{figure*}[t]
\centering
\vspace{-10pt}
\begin{subfigure}{1.9in}
    \centering
	\begin{tikzpicture}
		\node at (0,0) [scale=0.29]{\includegraphics{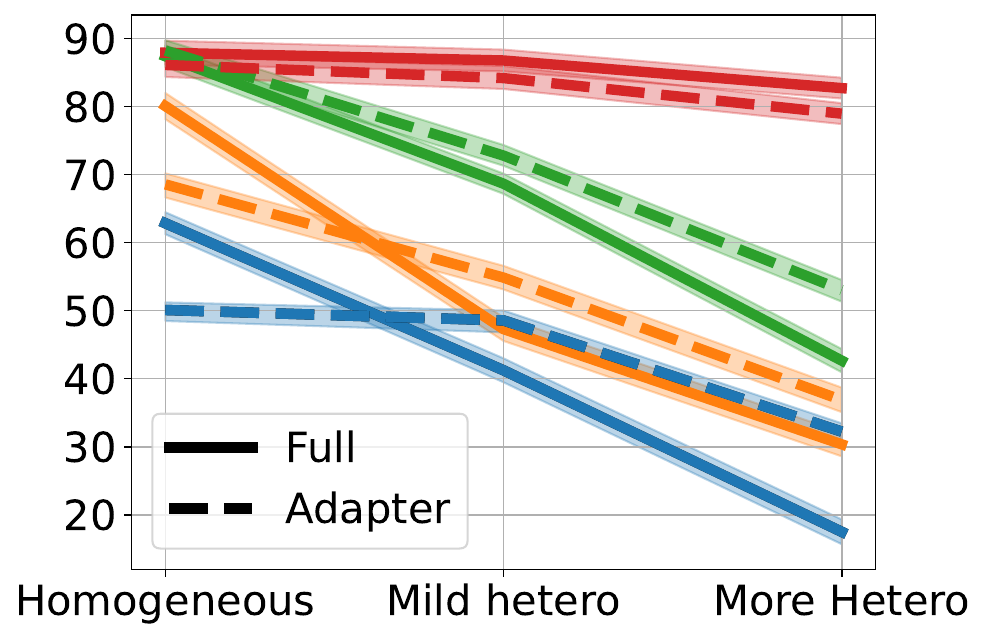}};
		\node at (-2.3 ,0) [scale=0.8,rotate=90] {Accuracy};
\end{tikzpicture}\vspace{-6pt}\caption{\small{Cifar10}}\label{fig:heter_cifar10_level}
\end{subfigure}
\begin{subfigure}{1.9in}
    \centering
	\begin{tikzpicture}
		\node at (0,0) [scale=0.29]{\includegraphics{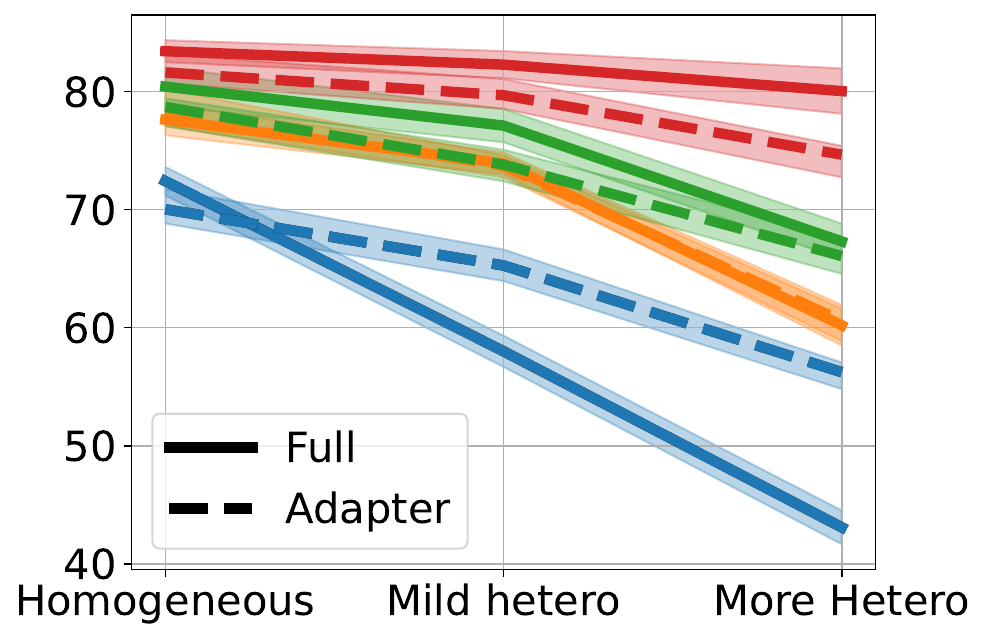}};
\end{tikzpicture}\vspace{-6pt}\caption{\small{Cifar100}}\label{fig:heter_cifar100_level}
\end{subfigure}
\hspace{-22pt}
\begin{subfigure}{0.7in}
    \centering
	\begin{tikzpicture}
		\node at (0,0) [scale=0.15]{\includegraphics{img/legend.pdf}};
\end{tikzpicture}\vspace{60pt}
\end{subfigure}
\vspace{-5pt}
\caption{\small{Test set accuracy under different levels of client data heterogeneity. Larger PTFs show consistently better performance, especially in more heterogeneous settings. Comparing the proportion of the same curve's descent from left to right, we observe that larger scale and modularity can enhance performance.}}
\label{fig:heter}
\end{figure*}

Similar results hold for clients with heterogeneous data distributions. Specifically, to evaluate performance on a heterogeneous data distribution, we use both simulated and real-world data heterogeneity. In Figure~\ref{fig:heter}, the simulated data heterogeneity setting involves clients with different class distributions, as defined in \S\ref{sec:expsetup}.
In Figure~\ref{fig:heter_acc}, 
the real-world data heterogeneity involves clients with both different class distributions and
different domains, such as each client having data relating to a particular celebrity in CelebA.
In all cases, larger PTFs outperform smaller PTFs.
A detailed discussion of the effects of data heterogeneity is deferred to \S\ref{sec:heterogeneous}.


\textbf{Larger PTFs narrow the local vs. federated training gap:}
The good performance of large PTFs raises the question of whether it is preferable to simply have clients store large PTFs locally and train them, without joint training through federated learning.
To study this, we conduct experiments to directly compare federated with purely local training.
Intuitively, federated learning should perform better since information is shared between clients, but larger PTFs may approach the performance of federated learning.
In Figs.~\ref{fig:cifar100_full_fedlocal},\ref{fig:cifar100_adapter_fedlocal}, we compare the performance of $\local$ and $\FedAvg$ for the \fullup and \modular training strategies.
The results show that $\local$ becomes increasingly competitive with $\FedAvg$ as the model scale increases (i.e., the gap between the solid and dashed lines is smaller for larger PTFs). For instance, in the case of the \modular training strategy with 16 samples per client  in Fig.~\ref{fig:cifar100_adapter_fedlocal}, the accuracy gap between the largest PTF (ViT-L, red line) for the \local and \FedAvg strategies is 8.10, while for the smallest PTF (ViT-T, blue line), the gap is 15.44.
Moreover, employing modules can help achieve better performance and narrow the gap between fully local and federated training (the gap between \local and \FedAvg is smaller in Fig.~\ref{fig:cifar100_adapter_fedlocal} than Fig.~\ref{fig:cifar100_full_fedlocal}).
In other words, if clients wish to completely avoid federated learning, large PTFs with modular updates that are trained on-device only can achieve reasonable performance.

We also obtain similar conclusions in the heterogeneous setting, as presented in Fig.~\ref{fig:fulllocal_heter}. Take CIFAR-100 with \fullup for example, the accuracy differences between $\local$ and $\FedAvg$ decreases for larger PTFs, with differences of [20.25, 12.65, 8.997, 7.725] from small to large ViT models (looking at Fig.~\ref{fig:cifar100_fully_localpersonalize} from left to right).

\begin{figure*}[t]
\centering
\vspace{-0.5cm}
\begin{subfigure}{1.2in}
    \centering
	\begin{tikzpicture}
		\node at (0,0) [scale=0.2]{\includegraphics{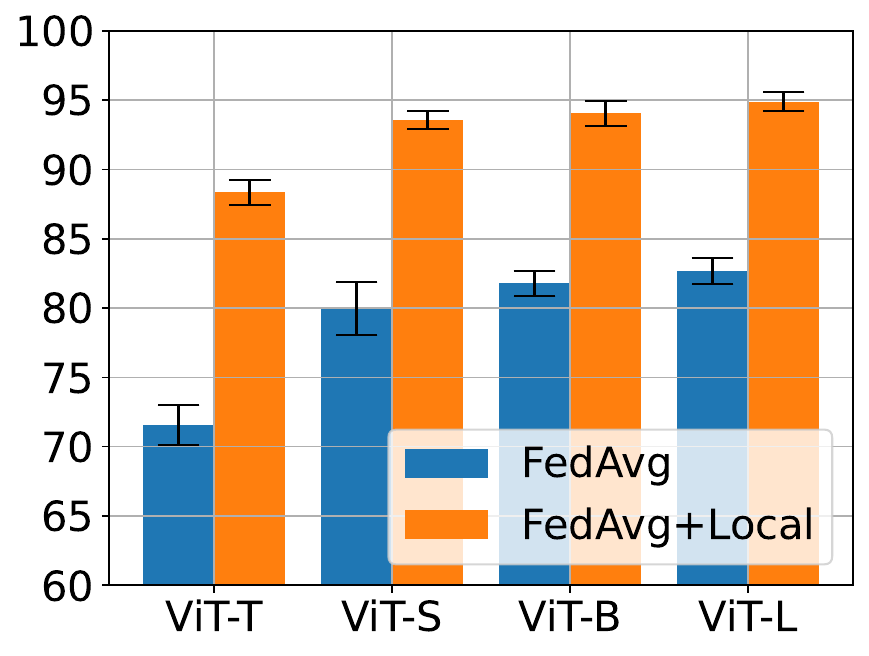}};
		\node at (-1.5 ,0) [scale=0.8,rotate=90] {Accuracy};
\end{tikzpicture}\vspace{-6pt}\caption{\small{CIFAR100,\\\modular}}\label{fig:cifar100_adapter_personalize}
\end{subfigure}
\begin{subfigure}{1.2in}
    \centering
	\begin{tikzpicture}
		\node at (0,0) [scale=0.2]{\includegraphics{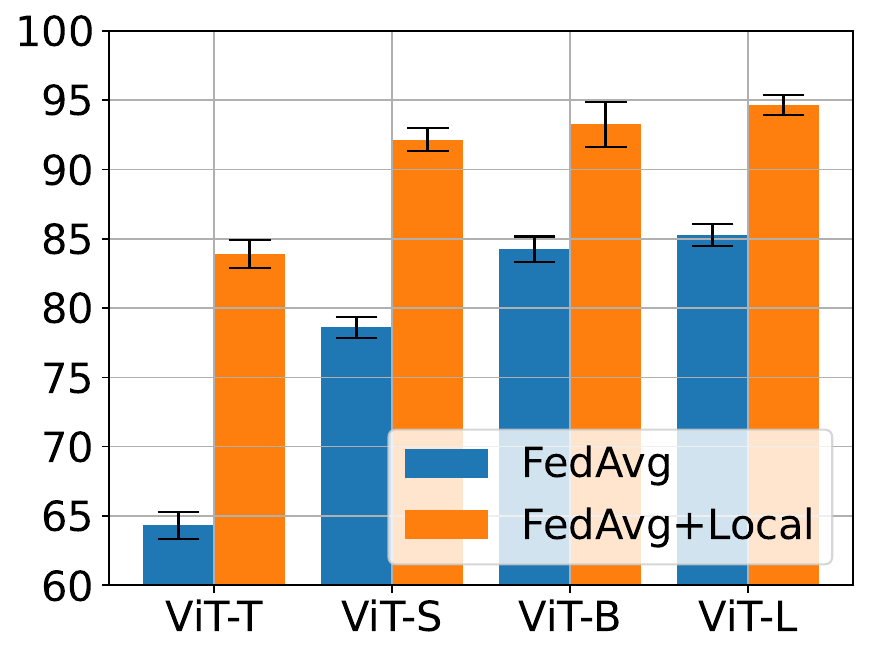}};
\end{tikzpicture}\vspace{-6pt}\caption{\small{CIFAR100,\\\fullup}}\label{fig:cifar100_fully_personalize}
\end{subfigure}
\begin{subfigure}{1.2in}
    \centering
	\begin{tikzpicture}
		\node at (0,0) [scale=0.2]{\includegraphics{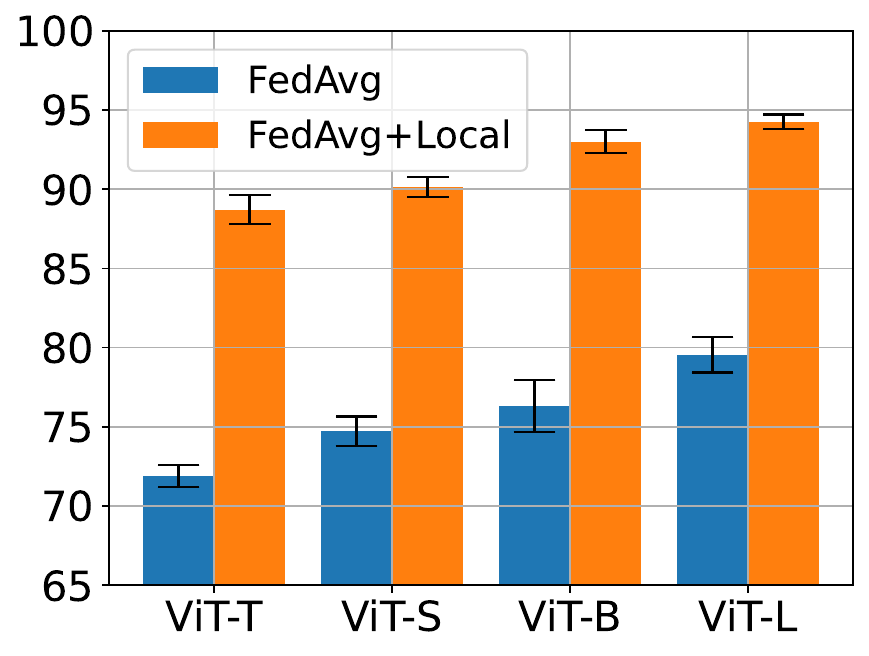}};
\end{tikzpicture}\vspace{-6pt}\caption{\small{CelebA,\\ \modular}}\label{fig:celeba_adapter_personalize}
\end{subfigure}
\begin{subfigure}{1.2in}
    \centering
	\begin{tikzpicture}
		\node at (0,0) [scale=0.2]{\includegraphics{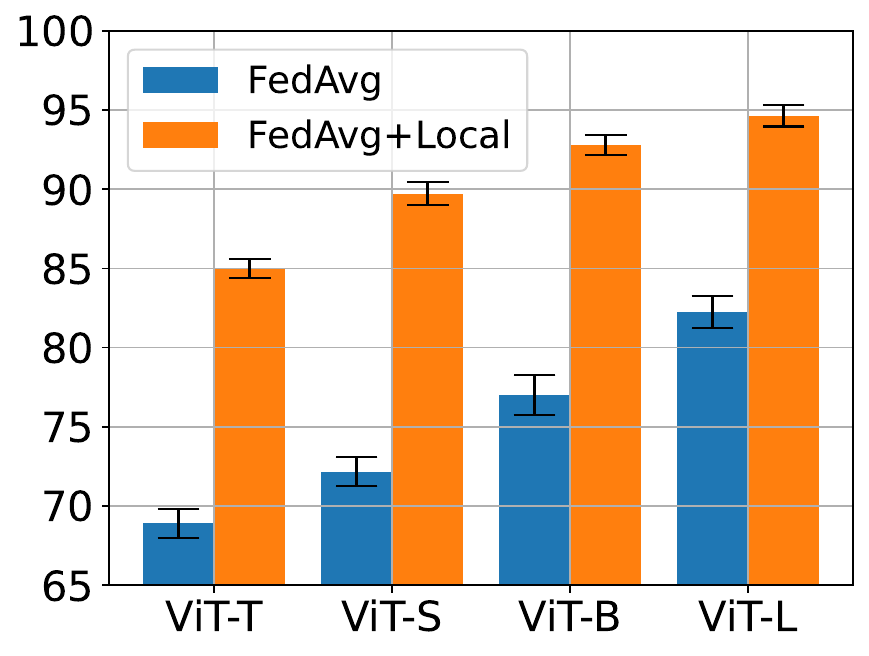}};
\end{tikzpicture}\vspace{-6pt}\caption{\small{CelebA,\\\fullup}}\label{fig:celeba_fully_personalize}
\end{subfigure}
\vspace{-5pt}
\caption{\small{Test set accuracy with (\Fedlocal) and without (\FedAvg) personalization. As the PTF scale increases, the gap between the two approaches diminishes.}}
\vspace{-10pt}
\label{fig:personalize_ornot}
\end{figure*}

\begin{figure*}[t]
\centering
\begin{subfigure}{2.2in}
    \centering
	\begin{tikzpicture}
		\node at (0,0) [scale=0.29]{\includegraphics{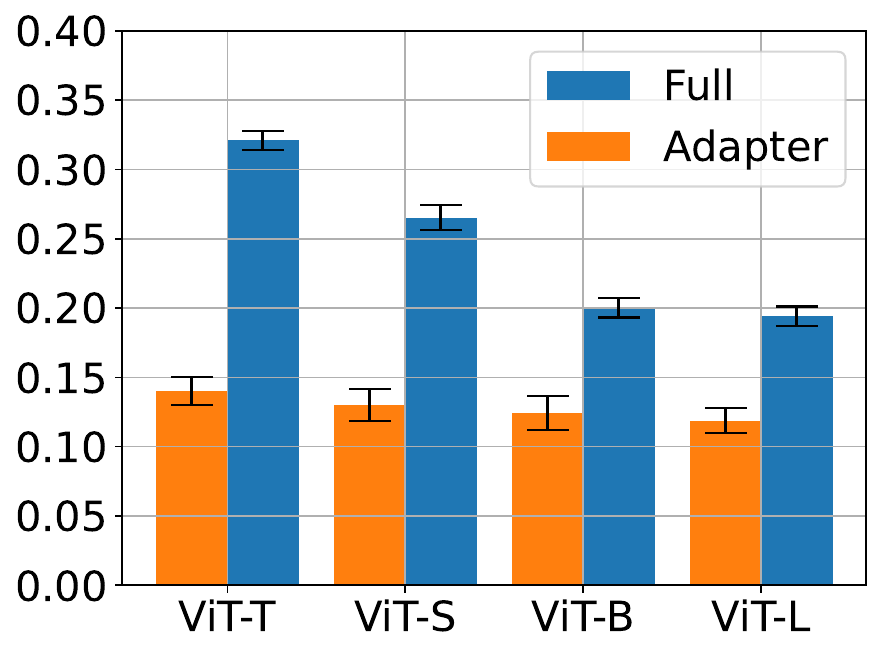}};
		\node at (-2.3 ,0) [scale=0.8,rotate=90] {Forgetting Ratio};
        \node at (0,-1.6) [scale=0.8] { }; 
\end{tikzpicture}\caption{\small{Global accuracy forgetting ratio}}\label{fig:forgetting_ratio}
\end{subfigure}
\hspace{10pt}
\begin{subfigure}{3in}
    \centering
	\begin{tikzpicture}
		\node at (0,0) [scale=0.29]{\includegraphics{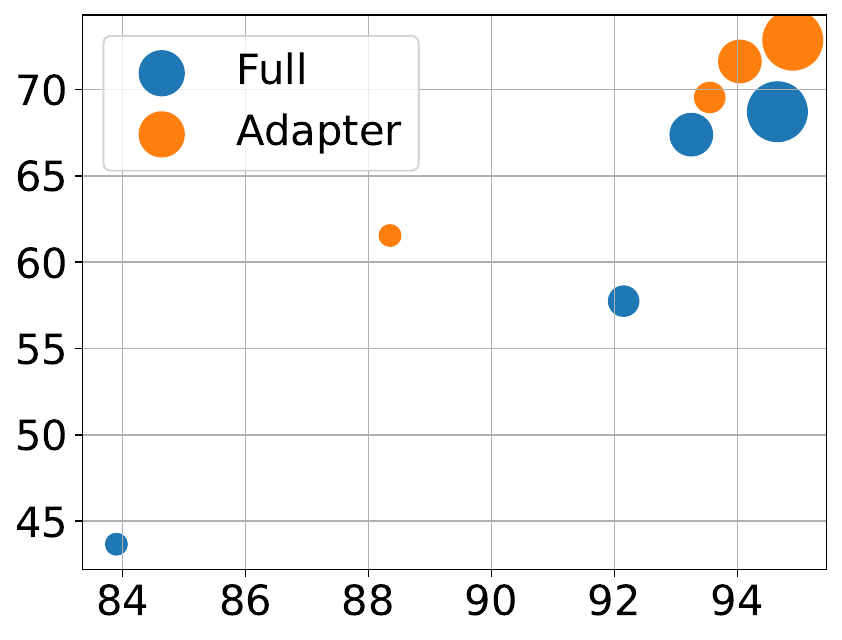}};
		\node at (-2.3 ,0) [scale=0.8,rotate=90] {Global Accuracy};
		\node at (0,-1.6) [scale=0.8] {Local Accuracy};
\end{tikzpicture}\vspace{-6pt}\caption{\small{Personalized model accuracy on global data vs.~local data}}\label{fig:forgetting_localgloabl}
\end{subfigure}
\caption{\small{Catastrophic forgetting. (a) Forgetting ratio (smaller value is better) vs model size and fine-tuning methods. \Modular and larger scale mitigates catastrophic forgetting. (b) $\Fedlocal$ accuracy on local (new class distribution, 20 personalized classes) and global (previous class distribution, 100 common classes) test datasets. The size of the circles corresponds to the size of the models from ViT-Tiny to ViT-Large.
Values towards the upper right corner are better. \Modular with large PTFs 
exhibit better retention of information from the previous class distribution (i.e.~post personalization).
}}
\label{fig:forget}
\end{figure*}
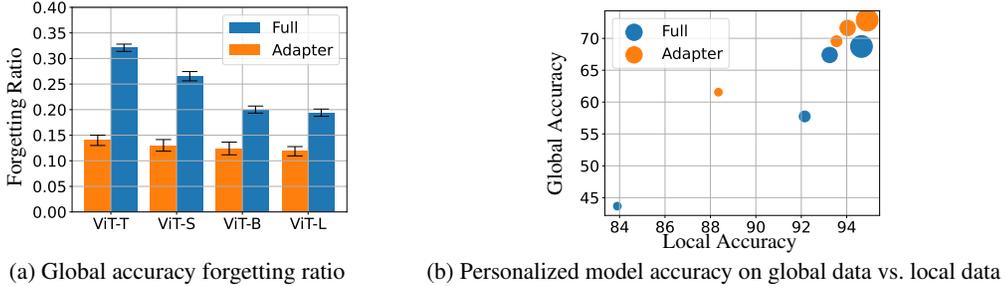

\textbf{Modular updates can outperform full updates:} 
In a previous study \cite{lester2021power}, it was demonstrated that with larger scale PTFs, the performance gap between full updates (\fullup) and prompt tuning could be reduced. In our experiments, displayed in Fig.~\ref{fig:cifar100_fed_fullmodule}, we observe similar findings in the context of federated learning, as the gap between \fullup (solid lines) and \modular (dashed lines) generally shrinks for larger PTFs.
This observation also extends to multiple modular methods, such as Adapter~\cite{houlsby2019parameterefficient}. 

However, federated learning introduces additional challenges related to heterogeneity and decoupled data, leading to more interesting findings. 
In particular, we find that modular approaches can actually outperform full updates in certain situations. This is particularly evident when using smaller PTFs and training with a very limited number of samples.
 For example in Figure~\ref{fig:cifar100_fed_fullmodule}, for a small number of training samples per client $(\leq 4)$, the \modular ViT-T outperformed \fullup.
 Similarly, with heterogeneous data distribution in Fig.~\ref{fig:heter_acc}, the ViT-T PTF sometimes have higher accuracy with the \modular training approach. The advantage of \modular is more pronounced when the data is even more heterogeneous, as depicted on the right half of Fig.~\ref{fig:heter}.  From these observations, we conclude that \fullup is more susceptible to issues introduced by federated learning, especially when using small-scale PTFs, and modular approaches can sometimes outperform them.

\vspace{-5pt}
\subsection{Heterogeneous client data distributions} \label{sec:heterogeneous}
In this section, our goal is to explore whether PTF scale and modularity can potentially improve robustness to heterogeneous data distributions across clients, bridge the gap between personalized and global models in FL, and mitigate catastrophic forgetting.
\textbf{Enhancing Robustness to Heterogeneous Distributions}:
In Fig.~\ref{fig:heter}, we plot the accuracy for \fullup and \modular training strategies, for varying amounts of data heterogeneity on the clients (as defined in \S\ref{sec:expsetup}).
The results show a notable decrease in test accuracy on heterogeneous data partitions when training smaller PTFs with full updates (solid blue curve), particularly in the highly heterogeneous setting. Employing larger PTFs or modular update maintains accuracy even under significant heterogeneity. Larger PTFs consistently outperform, irrespective of heterogeneity level or fine-tuning method. If PTFs are not sufficiently large, performance plummets as heterogeneity escalates (e.g., the solid blue curve). In contrast, modular update can enhance performance.

We assess performance retention by calculating the accuracy decrease proportion $\frac{\acc_{Hom}-\acc_{Hetero}}{\acc_{Hom}}$. As heterogeneity rise, larger backbones and modular update substantially lower this proportion. In CIFAR-100, transitioning from homogeneous to more heterogeneous data results in performance decrease proportions of 8.5\% with ViT-L and \modular, 19.7\% with ViT-T and \modular, and 40.5\% with ViT-T and \fullup. Hence, larger PTFs exhibit greater robustness to heterogeneous data, and modular update can bolster smaller models' robustness.

\textbf{Bridging the Personalization Gap}:
Personalized training, commonly used to adapt models to individual clients in federated learning, can pose challenges due to each client's limited sample sizes and the associated computational cost. We next explore whether PTFs and modularity can help reduce the disparity between personalized training and the average global model. As shown in Figure~\ref{fig:personalize_ornot}, the disparity shrinks as the scale of the PTFs grows, for different datasets and update strategies. Full update tends to widen this gap, especially with smaller backbones, in contrast to the modular update. This suggests that employing larger PTFs and modular update could mitigate the necessity for computationally intensive personalized training.


\textbf{Mitigating Catastrophic Forgetting}:
We examine if larger scale and modularity can alleviate catastrophic forgetting, as depicted in Fig.~\ref{fig:forget}. The model's performance is compared pre- and post- personalization to induce forgetting. Initially, the model is trained on a global dataset, followed by personalization by training on a client-specific local dataset with fewer classes. Maintaining performance on the full set of 100 classes alongside improving accuracy on the local classes indicates better forgetting resistance.
Fig.~\ref{fig:forgetting_ratio} shows the forgetting ratio ($\frac{Acc_{\FedAvg}-Acc_{\Fedlocal}}{Acc_{\FedAvg}}$, smaller is better) for different PTF sizes.
The results demonstrate that \modular significantly reduces the forgetting ratio, with this ratio decreasing as the PTF scale increases.
Fig.~\ref{fig:forgetting_localgloabl} plots the global accuracy vs. the local accuracy for different PTF models and update strategies (\fullup, \modular).
The results show that larger PTFs and modularity enable personalized models to simultaneously achieve higher global and local accuracy, effectively mitigating catastrophic forgetting. 


\subsection{Reducing Federated Communication Cost with Modular Updates} 
\label{sec:exp_efficiency}

\begin{figure}
\centering
\begin{subfigure}{1.7in}
    \centering
	\begin{tikzpicture}
		\node at (0,0) [scale=0.29]{\includegraphics{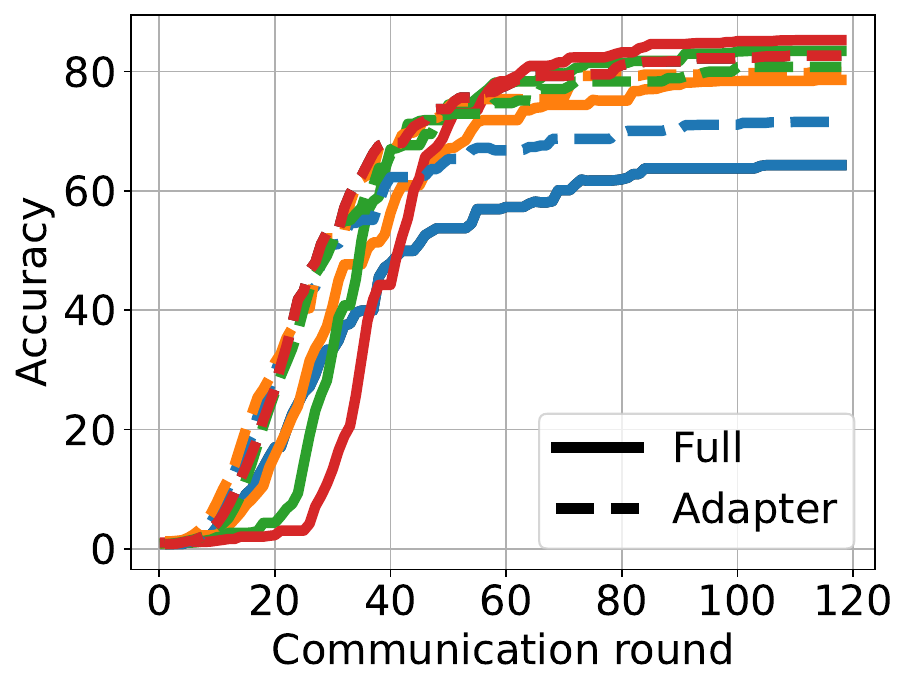}};
\end{tikzpicture}\label{fig:comm_round}
\end{subfigure}
\begin{subfigure}{1.7in}
    \centering
	\begin{tikzpicture}
		\node at (0,0) [scale=0.29]{\includegraphics{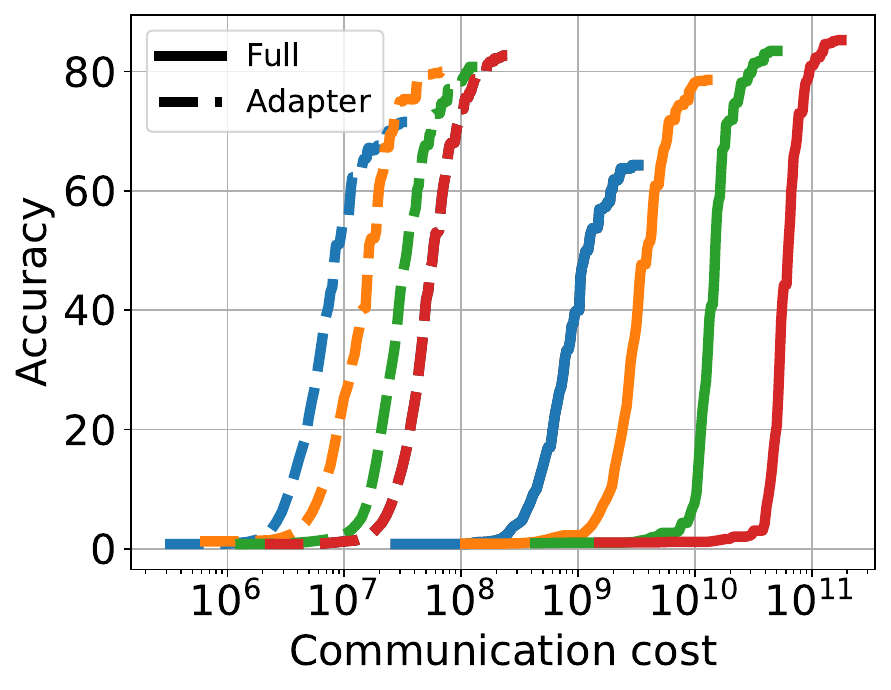}};
\end{tikzpicture}\label{fig:comm_cost}
\end{subfigure}
\begin{subfigure}{0.7in}
    \centering
	\begin{tikzpicture}
		\node at (0,0) [scale=0.15]{\includegraphics{img/legend.pdf}};
\end{tikzpicture}\vspace{50pt}
\end{subfigure}
\caption{\small{Communication cost is greatly decreased with \modular compared to \fullup. }}
\label{fig:comm}
\end{figure}

\begin{table*}[t]
\centering
\vspace{-4pt}
\begin{tabular}{lllll}
        & ViT-T & ViT-S & ViT-B & ViT-L \\ \hline
\Fullup    & 5,543,716 & 21,704,164    &   85,875,556    &   303,404,132 \\
\Modular  &   58,564    &     116,932  &  233,668     &    417,984   \\     
\end{tabular}\vspace{-5pt}\caption{\small{Number of parameters for different PTF scales.}}\label{tab:num_param}
\vspace{-10pt}
\end{table*}
\begin{figure}[h]
\centering
\begin{subfigure}{1.7in}
    \centering
	\begin{tikzpicture}
		\node at (0,0) [scale=0.29]{\includegraphics{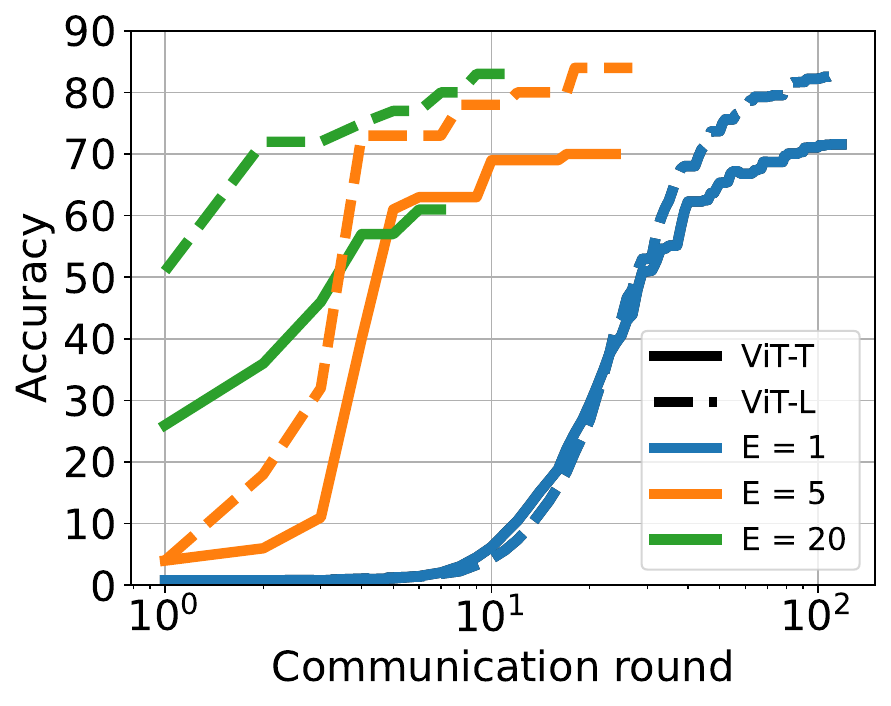}};
\end{tikzpicture}\vspace{-4pt}\caption{CIFAR100}\label{fig:ablation_localepoch}
\end{subfigure}
\begin{subfigure}{1.7in}
    \centering
	\begin{tikzpicture}
		\node at (0,0) [scale=0.29]{\includegraphics{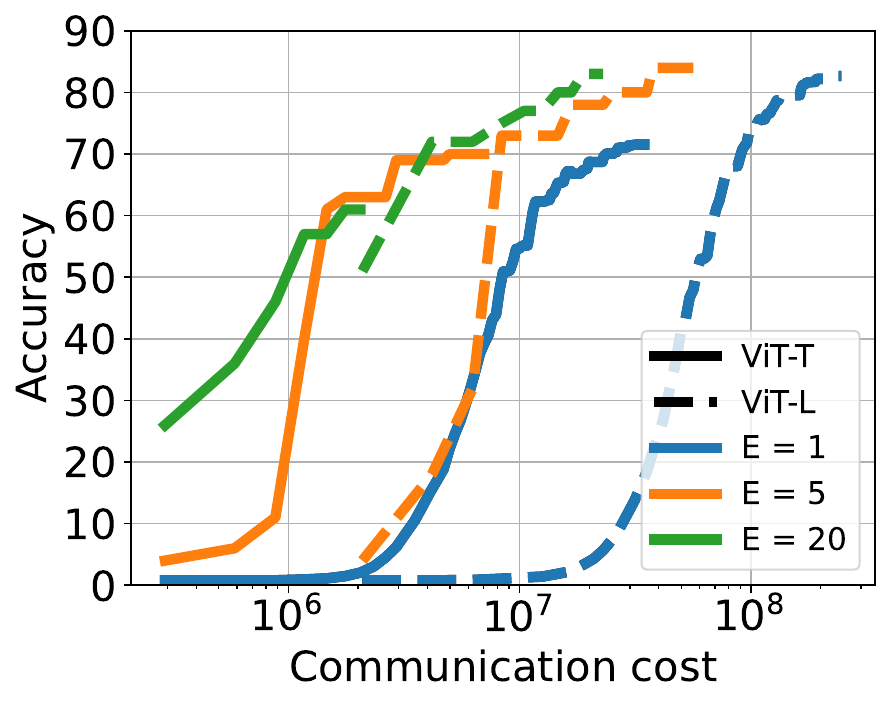}};
\end{tikzpicture}\vspace{-4pt}\caption{CIFAR100}\label{fig:ablation_localepoch_commcost}
\end{subfigure}\\
\begin{subfigure}{1.7in}
    \centering
	\begin{tikzpicture}
		\node at (0,0) [scale=0.29]{\includegraphics{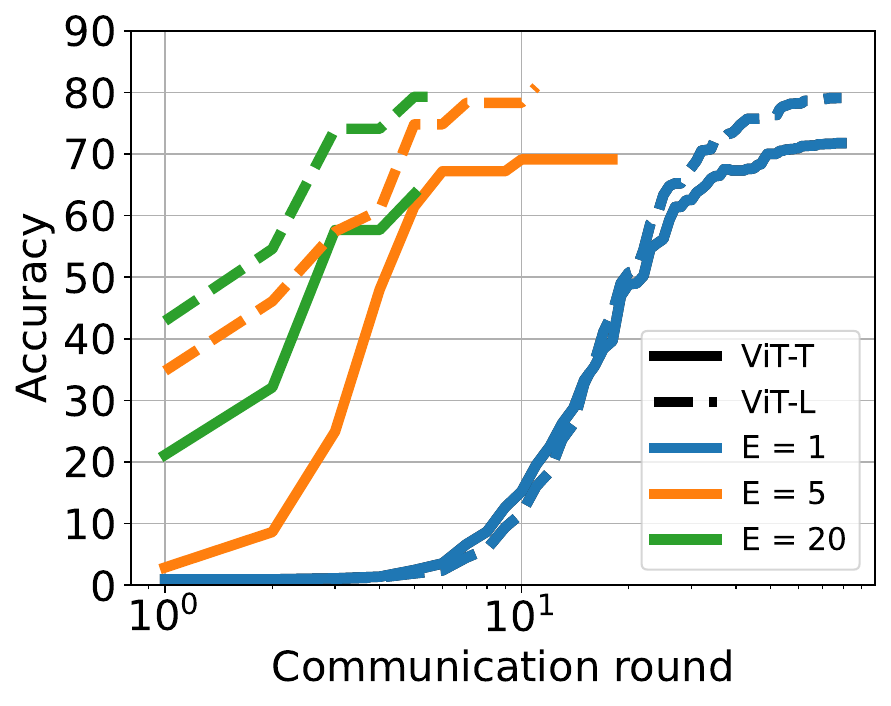}};
\end{tikzpicture}\vspace{-4pt}\caption{CelebA}\label{fig:ablation_localepoch_celeba}
\end{subfigure}
\begin{subfigure}{1.7in}
    \centering
	\begin{tikzpicture}
		\node at (0,0) [scale=0.29]{\includegraphics{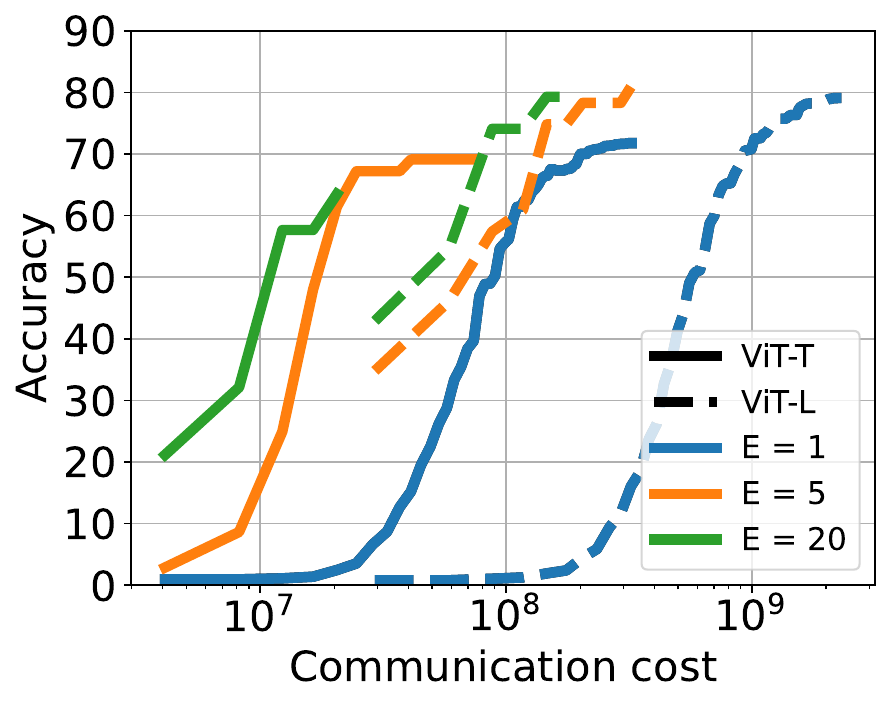}};
\end{tikzpicture}\vspace{-4pt}\caption{CelebA}\label{fig:ablation_localepoch_commcost_celeba}
\end{subfigure}
\caption{We conducted experiments comparing two different scales of PTFs, ViT-large (ViT-L, dashed line) and ViT-tiny (ViT-T, solid line). The color represents the number of local training epochs ($E$) used in the experiments. All experiments were performed using the \modular approach with Adapter. In order to showcase the convergence speed, we employed early stopping when the training reached convergence.}
\label{fig:comm_localepoch}
\end{figure}

\begin{table*}[t]
\centering
\begin{tabular}{lllll}
   &  Dataset   & \# of clients & \# of samples per client & Data partition \\ \hline
Task0& CIFAR100    & 20 &     100       &  Mild hetero  \\
Task1& CIFAR10    & 20 &     100       &  Mild hetero  \\ 
Task2& CelebA   & 787 &   $\leq$8      &  Based on celebrity identity \\ 
Task3& FEMNIST   & 532 &   $\leq$120      & Based on character writer  
\end{tabular}\vspace{-5pt}\caption{\small{Details of data partitioning for multi-task learning}.}\label{tab:multitask}
\vspace{-10pt}
\end{table*}
Our work aims to reduce communication cost while preserving accuracy, an objective intuitively achieved through modules that require fewer parameters for training than updating the entire model. Table~\ref{tab:num_param} shows the number of transmitted parameters $\numpara$, which is much smaller for \modular compared to \fullup across all ViT architectures. We maintain a consistent learning rate for a fair comparison.
The experiments are conducted with mild heterogeneous CIFAR-100.

\textbf{Modularity Decreases Communication Rounds:} 
We compare the number of FL communication rounds required by \modular and \fullup, plotted in Fig~\ref{fig:comm} (left). The \modular approach  (dashed lines) outperforms \fullup(solid line) during initial training stages and achieves target accuracy in fewer epochs 37.25 ($\pm 1.08$) on average for \modular, versus 47.25 ($\pm 1.48$) for \fullup. Contrary to previous studies \cite{huang2022fpt,ding2022delta}, we find that modular updates typically converge faster in the federated setting.


\textbf{Modularity significantly reduces communication cost, by over 100x: } 
The communication cost can be defined as $T \times M \times P$, where $T$ is the number of communication rounds, $M$ is the number of clients per round, and $P$ is the number of transmitted parameters.
We plot the accuracy as a function of communication cost in Fig~\ref{fig:comm} (right). \modular significantly reduces the number of transmitted parameters compared to \fullup, for all model sizes.
This improvement in efficiency helps address the communication bottleneck commonly associated with federated learning.

\textbf{Large PTFs allow more local epochs: } 
Previous research has recommended using large local training epochs ($E$) on homogeneous data to reduce communication costs. However, for highly heterogeneous data partitions, it is suggested to use a small $E$ ($E \leq 5$ for $\FedAvg$). This is because larger E may result in a decline in final performance on heterogeneous data partitions. In our study, we demonstrate that larger scales of PTFs can enable larger local training epochs even with heterogeneous data partitions. The results are presented in Fig.~\ref{fig:comm_localepoch}.

In Fig.~\ref{fig:ablation_localepoch}, we compare the communication rounds associated with different scales of PTFs and varying numbers of local training epochs ($E$). By comparing different $E$ (colors), it is evident that using larger local training epochs ($E$) can significantly accelerate convergence. For fine-tuning with small PTFs (solid line), it was observed that larger $E$ negatively impacted the performance, which aligns with previous findings. However, the dashed line demonstrates that larger-scale PTFs can maintain or even improve performance when larger $E$ values are used. 

In Figure~\ref{fig:ablation_localepoch_commcost}, we compare the communication cost. As shown in Figure~\ref{fig:comm}, larger-scale PTFs generally exhibit a higher communication cost due to their larger number of trainable parameters. However, our findings in Figure~\ref{fig:ablation_localepoch_commcost} reveal that by simply increasing the number of local training epochs ($E$), larger-scale PTFs can achieve comparable or even better performance than smaller-scale PTFs within a fixed communication cost budget. For instance, for CIFAR100 dataset, when the communication cost is limited to $10^7$, ViT-T with $E=1$ achieves an accuracy of 54.61 with 33 communication rounds. In contrast, ViT-L with $E=20$ achieves a higher accuracy of 75.04 with only 3 communication rounds. 

\begin{figure*}[t]
\centering
\begin{subfigure}{1.7in}
    \centering
	\begin{tikzpicture}
		\node at (0,0) [scale=0.29]{\includegraphics{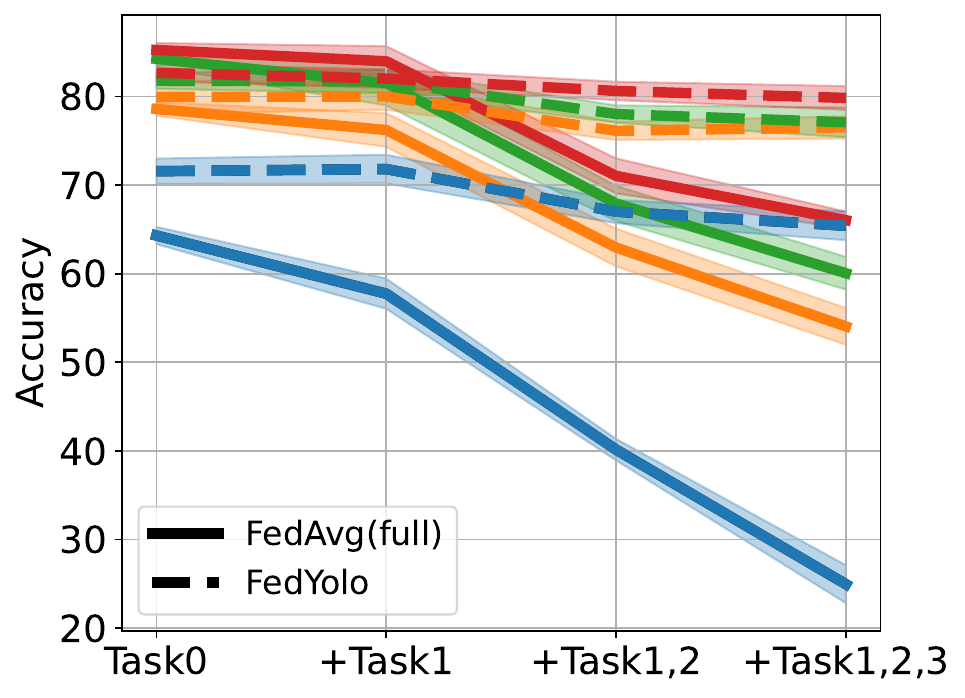}};
\end{tikzpicture}\vspace{-6pt}\caption{\small{Federated setting}}\label{fig:joint_cifar100}
\end{subfigure}
\hspace{-8pt}
\begin{subfigure}{0.7in}
    \centering
	\begin{tikzpicture}
		\node at (0,0) [scale=0.15]{\includegraphics{img/legend.pdf}};
\end{tikzpicture}\vspace{65pt}
\end{subfigure}
\begin{subfigure}{1.7in}
    \centering
	\begin{tikzpicture}
		\node at (0,0) [scale=0.29]{\includegraphics{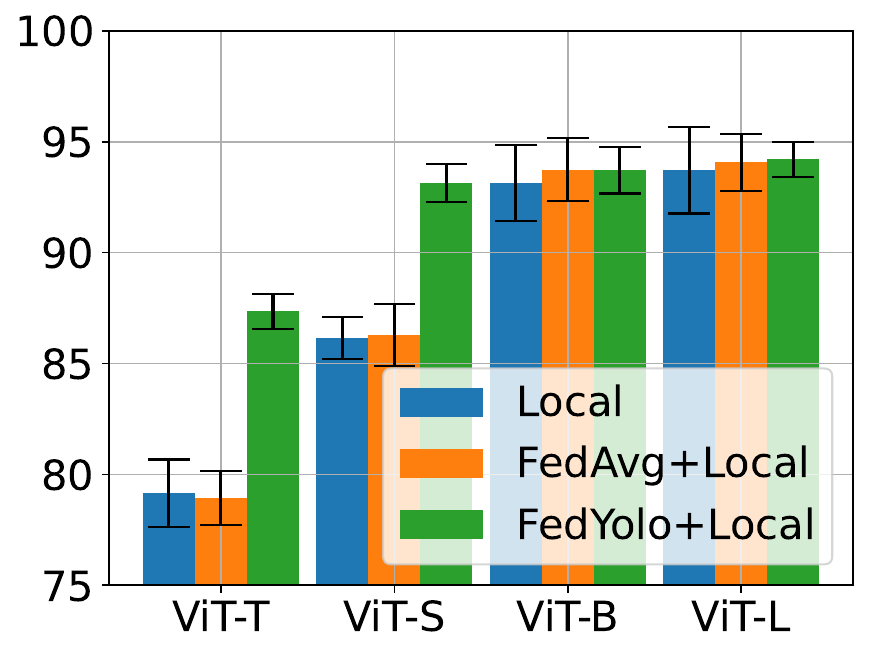}};
\end{tikzpicture}\vspace{-6pt}\caption{\small{Federated + Local}}\label{fig:joint_cifar100_personalize}
\end{subfigure}
\vspace{-5pt}
\caption{\small{Task 0 (CIFAR-100) performance of multi-task federated learning as we incorporate more tasks to the problem (y-axis). (a) Global accuracy, (b) Local accuracy with adaptation. In either scenario, \yolo, with its task-specific modules, outperforms conventional \FedAvg. 
}}
\vspace{-10pt}
\label{fig:multitask}
\end{figure*}
\begin{algorithm}[ht]\begin{multicols}{2}\small
\caption{\yolo}\label{alg:fedyolo}
\vspace{-10pt}
\textbf{Parameters:} 
Client set $\mathcal{C}$; \# of rounds $T$; \# of local epochs $E$; \# of tasks $K$; \# of clients per round $M$;

\vspace{2pt}
PTF parameters \darkred{$\mathcal{W}_{\texttt{frozen}}$}; trainable parameters \darkred{$\mathcal{W}_{\texttt{train}}^{k}$} for task $k$ (containing task-specific head and module);

\vspace{2pt}
Local dataset $\mathcal{D}_m$ of client $m$.
\vspace{2pt}
\begin{algorithmic}[1]
\State Load and freeze PTF $\mathcal{W}_{\texttt{frozen}}$ on each client
\For {each communication round $t=1$ to $T$}
    \State $\mathcal{C}^t \leftarrow$ (randomly sample $M$ clients from $\mathcal{C}$ )
    \For {each client $m \in \mathcal{C}^t$ \textbf{in parallel }}  
    \State $k$ $\leftarrow$ task ID of client $m$
    \State Load $\mathcal{W}_{\texttt{train}}^{t,k}$ to the client
    \State \parbox[t]{\dimexpr\linewidth-\algorithmicindent\relax}{\raggedright$\mathcal{W}_{\texttt{train}}^{t+1,m,k} \leftarrow \textsc{LocalTuning}(m, \mathcal{W}_{\texttt{train}}^{t,k})$}
    \State Send client parameters $\mathcal{W}_{\texttt{train}}^{t+1,m,k}$ to server
    \EndFor 
     \For{task $k=1$ to $K$}
     \State $\mathcal{C}^{t,k}$ $\leftarrow$ clients in $\mathcal{C}^t$ with task $k$
     \State $\mathcal{W}_{\texttt{train}}^{t+1,k}$ $\leftarrow$ \texttt{Average}$(\{\mathcal{W}_{\texttt{train}}^{t+1,m,k}\}_{m \in \mathcal{C}^{t,k}}$)
     \EndFor
\EndFor\\
\Function{LocalTuning}{$m$, $\mathcal{W}_{\texttt{train}}^{t,k}$} 
    \State \parbox[t]{\dimexpr\linewidth-\algorithmicindent\relax}{\raggedright$\mathcal{W}^t$ $\leftarrow$ (assemble task-specific $\mathcal{W}_{\texttt{train}}^{t,k}$ and $\mathcal{W}_{\texttt{frozen}}$)
    } 
    \For {epoch $e=1$ to $E$}
            \State $\mathcal{W}_{\texttt{train}}^{t+1,m,k}$ $\leftarrow$ train $\mathcal{W}_{\texttt{train}}^{t,k}$ on dataset $\mathcal{D}_m$
    \EndFor
    \State Send $\mathcal{W}_{\texttt{train}}^{t+1,m,k}$ to the server
\EndFunction
\end{algorithmic}\end{multicols}
\vspace{-5pt}
\end{algorithm}

\section{Federated Multitask Learning via FedYolo} \vspace{-5pt}
\label{sec:exp_multitask}

 Traditional $\FedAvg$ entails high communication costs and vulnerability to heterogeneity due to shared full PTF parameters.
 The experiments within Section~\ref{sec:results} have demonstrated the potential of large-scale PTFs and modularity to reduce communication costs and boost robustness, making them promising for multi-task federated learning. 
 Based on these findings, we propose $\yolo$ as a multi-task federated learning algorithm.
 Our $\yolo$ method, described in Algorithm~\ref{alg:fedyolo} and illustrated in Figure~\ref{fig:yolo_fig} is conceptually a simple and intuitive idea: it assigns a unique module to each task where the modules are plugged into a single frozen PTF. This PTF is loaded once at the start of the training, equipping the client with a backbone architecture. The task-specific modules are then updated and communicated with minimal communication cost.

Note that, in Algorithm~\ref{alg:fedyolo}, each client trains and sends modules for their own tasks. This might potentially suffer from privacy leak as the server will know which client has what task/module. An alternate approach is letting clients send a vector of length $K \times$ single module size, where most of the entries are zero and only the tasks at hand have non-zero entries. This can be combined with secure aggregation techniques \cite{fereidooni2021safelearn,mansouri2023sok} to ensure that the server will not learn which clients contributed to a particular task.


To evaluate $\yolo$, we train clients on multiple tasks simultaneously, including image classification on CIFAR-10, CIFAR-100, CelebA, and FEMNIST datasets, where each client is assigned to one task. The task assignments and data partitioning details are in Table~\ref{tab:multitask}.
A \FedAvg baseline with \fullup is also trained on the same tasks. We display the evolving accuracy of Task 0.
The results in Figure~\ref{fig:multitask} show $\yolo$ (dashed line) consistently outperforming conventional \FedAvg (solid line), particularly with more tasks and for smaller PTFs.
To examine the impact of personalization, we conduct another experiment where we add local training for clients after the federated training is complete (see \S\ref{sec:expsetup} for details).
The results in Fig.~\ref{fig:joint_cifar100_personalize} show that $\yolo$ also surpasses $\FedAvg$ and $\local$ with personalized models in terms of accuracy, especially with smaller models. The performance gap narrows with larger models, supporting larger PTFs' role in balancing local and federated training. With larger PTFs, users can exclusively train locally with similar performance, valuable where data privacy is vital. 
Across PTF sizes, the near-identical performance of $\local$ and $\Fedlocal$ implies standard \FedAvg's limited impact on the global model's generalization ability, whereas $\yolo$ provides clear improvements by avoiding interference across distinct tasks.

\vspace{-5pt}\section{Conclusions, Limitations, and Future Directions}\vspace{-5pt}
\label{sec:conclusion}

We have shown that the scale and modularity of pretrained transformers have the potential to address important bottlenecks in federated learning and proposed \yolo as an effective multitask learning scheme. An important consideration we have not explored in this work is the computational cost associated with deploying PTFs as this is a clear drawback of deploying large models. We note that one can mitigate the computational cost by employing compression/distillation techniques, early-exit strategies (where computation terminates at initial layers), or offloading methods \cite{wang2019edge,feng2022computation} which deserve a separate investigation. Our work also motivates new avenues for future research: Ideally, PTFs will also prove helpful in other challenging settings, such as, when clients have unlabeled data or non-stationary distributions. Additionally, it might be interesting to devise more sophisticated methods of \yolo by assigning shared modules across tasks or clients and by searching for optimal module placements within PTF.



\bibliographystyle{plain}
\bibliography{fedyolo}


\newpage

\appendix
\section{Organization of the Appendix}
In Section \ref{sec:app_exp}, we add additional experiments. Specifically, we make the following observations: 
\begin{itemize}
  \item Large PTFs allow for using more local epochs without sacrificing accuracy. This reduces the number of communication rounds in federated learning.
  \item We provide evaluations for FedProx which is a state-of-the-art optimization-based federated learning method for heterogeneity. In line with main submission, \yolo outperforms FedProx with full-updates in multitask settings.
  \item In main submission, we only compared full-update and modular-update. An alternative is only tuning the classifier head i.e.~the final layer(s). We find that modular-update achieves superior performance compared to only head-tuning under similar number of trainable parameters.
\item Our main empirical findings generalize well across module types (LoRa, Adapted, prompt-tuning).
\item Larger PTF retain its benefits over smaller PTF even if we use the same module size (i.e.~equalizing the number of trainable parameters). We conducted this experiment because in the main body of the paper, we used the default module sizes which are proportional to the embedding dimension, thus, larger PTFs were using larger modules.
\end{itemize}
In Section \ref{sec:app_exp_deatil}, we provide further experiment details. In Section \ref{sec:explanation}, we propose a representation-based explanation for the improved performance observed with modularity and larger-scale PTFs.
\begin{figure}[!b]
\centering
\vspace{-10pt}
\begin{subfigure}{1.8in}
    \centering
	\begin{tikzpicture}
		\node at (0,0) [scale=0.29]{\includegraphics{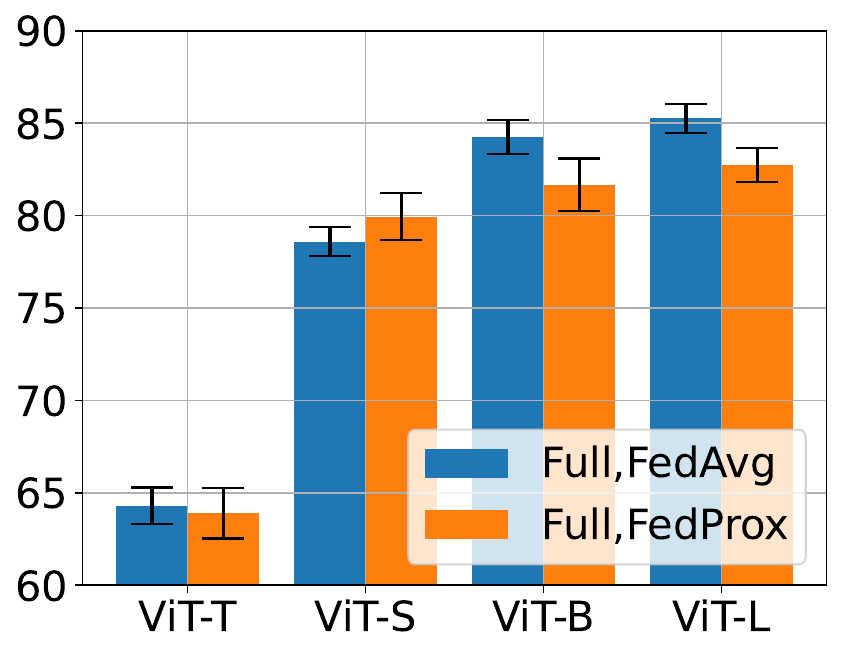}};
  		\node at (-2.1 ,0) [scale=0.8,rotate=90] {Accuracy};
\end{tikzpicture}\vspace{-4pt}\caption{\Fullup: \\\FedAvg v.s. \texttt{FedProx}\xspace}\label{fig:fedprox_fully}
\end{subfigure}
\begin{subfigure}{1.7in}
    \centering
	\begin{tikzpicture}
		\node at (0,0) [scale=0.29]{\includegraphics{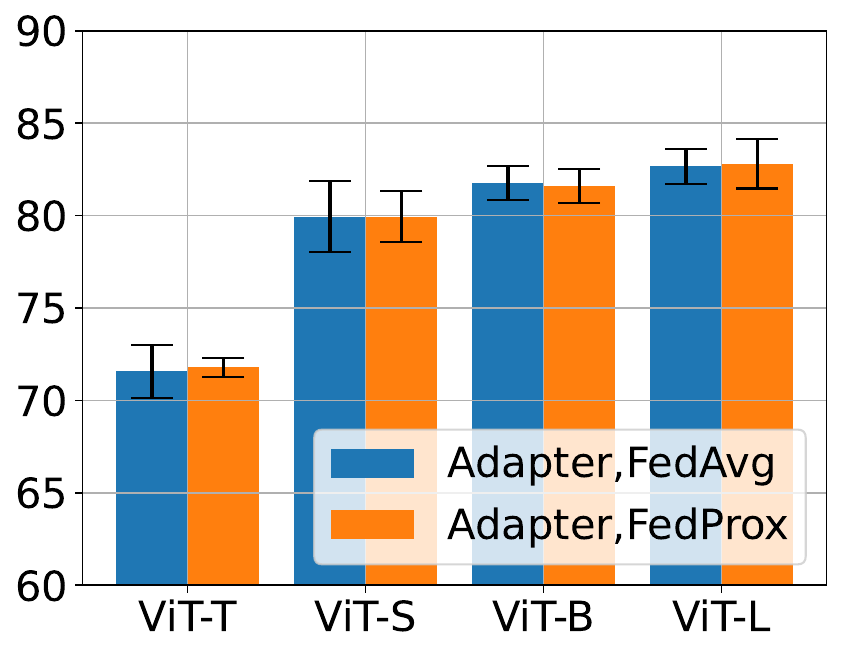}};
\end{tikzpicture}\vspace{-4pt}\caption{\Modular: \\\FedAvg v.s. $\texttt{FedProx}\xspace$}\label{fig:fedprox_adapter}
\end{subfigure}
\begin{subfigure}{1.7in}
    \centering
	\begin{tikzpicture}
		\node at (0,0) [scale=0.29]{\includegraphics{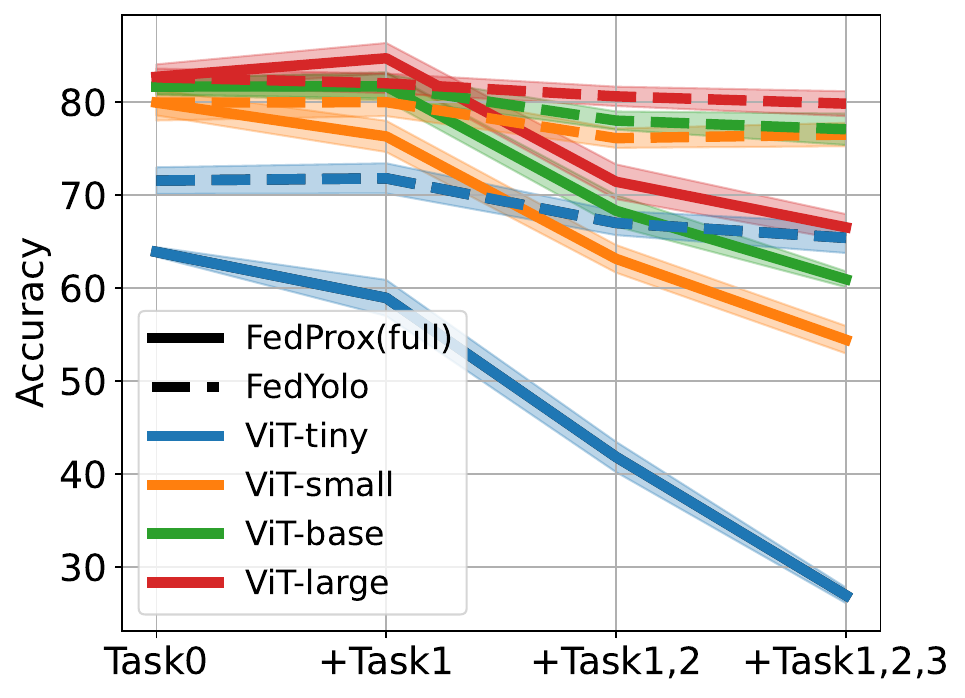}};
\end{tikzpicture}\vspace{-4pt}\caption{Federated Multi-task learning}\label{fig:fedprox_multitask_adapter}
\end{subfigure}
\begin{subfigure}{1.8in}
    \centering
	\begin{tikzpicture}
		\node at (0,0) [scale=0.29]{\includegraphics{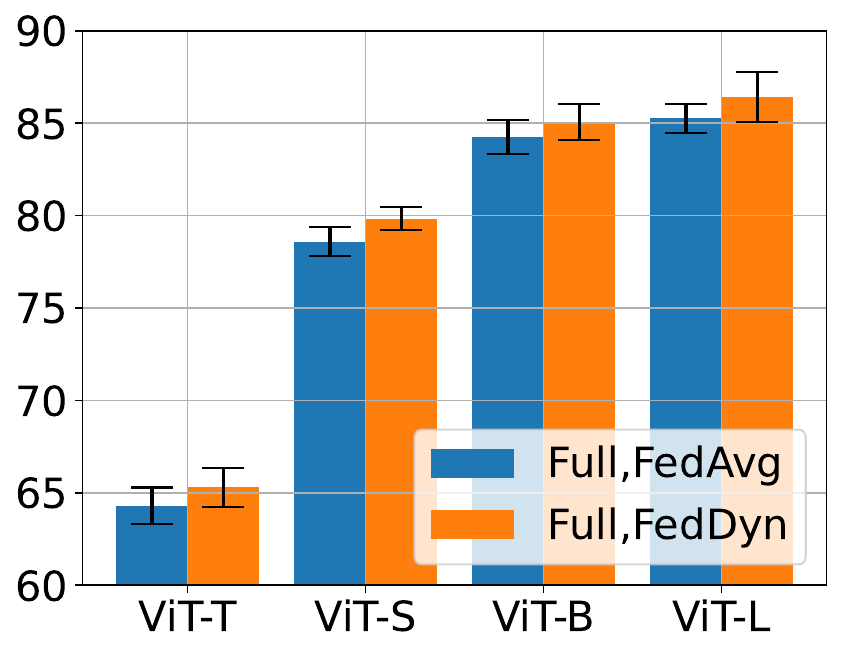}};
  		\node at (-2.1 ,0) [scale=0.8,rotate=90] {Accuracy};
\end{tikzpicture}\vspace{-4pt}\caption{\Fullup: \\\FedAvg v.s. $\texttt{FedDyn}\xspace$}\label{fig:feddyn_fully}
\end{subfigure}
\begin{subfigure}{1.7in}
    \centering
	\begin{tikzpicture}
		\node at (0,0) [scale=0.29]{\includegraphics{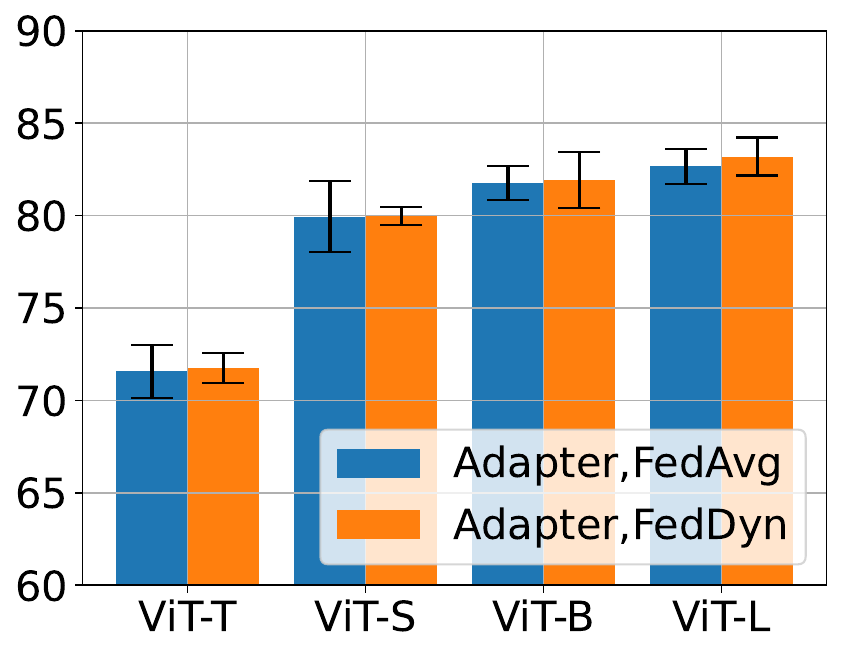}};
\end{tikzpicture}\vspace{-4pt}\caption{\Modular: \\\FedAvg v.s. $\texttt{FedDyn}\xspace$}\label{fig:feddyn_adapter}
\end{subfigure}
\begin{subfigure}{1.7in}
    \centering
	\begin{tikzpicture}
		\node at (0,0) [scale=0.29]{\includegraphics{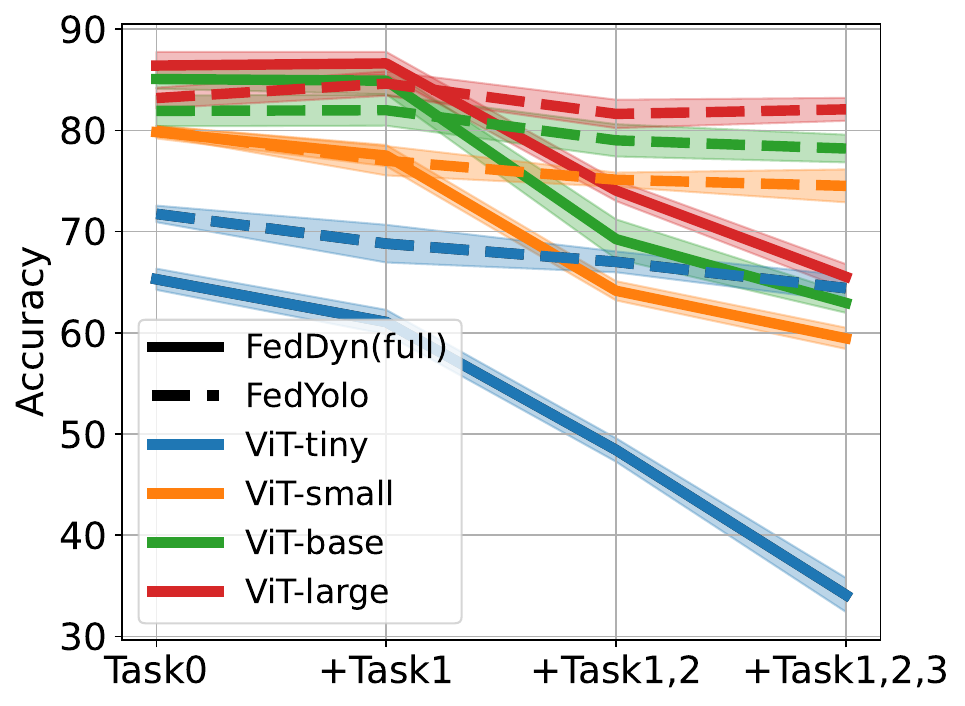}};
\end{tikzpicture}\vspace{-4pt}\caption{Federated Multi-task learning}\label{fig:feddyn_multitask_adapter}
\end{subfigure}
\caption{Experiment results using state-of-the-art optimization-based federated learning method $\texttt{FedProx}$ and $\texttt{FedDyn}$, instead of \FedAvg. (a,b) display the model performance with a heterogeneous distribution for CIFAR-100, following the same setting as in Figure \ref{fig:heter_cifar100}. We compare the performance between 
$\texttt{FedProx}$ (orange) and \FedAvg(blue) for different update strategies (\fullup and \modular). The results show that $\texttt{FedProx}$ does not demonstrate notable improvement. In(c), we present the model performance in the context of federated multi-task learning. Our proposed method, $\yolo$, consistently outperforms  $\texttt{FedProx}$ with \fullup. Similar results were obtained for $\texttt{FedDyn}$ in figures (d,e,f).}
\vspace{-4pt}
\label{fig:prox}
\end{figure}

\section{Additional Experiments}\label{sec:app_exp}

\subsection{Comparisons to Existing FL Methods}
We also compare our proposed method to the state-of-the-art optimization-based federated learning method $\texttt{FedProx}$ \cite{li2020federated} and $\texttt{FedDyn}$\cite{acar2021federated}. $\texttt{FedProx}$ uses a proximal term in the local objective function to mitigate weight divergence issues. We keep all the hyperparameters and set the penalty constant $\mu$ in the proximal term of $\texttt{FedProx}$ to 0.1. We tune the hyperparameter $\mu$ using ViT-B and \modular with a grid search approach and then apply the same value to all the other scales of PTFs and update strategy (\fullup). The results are shown in Fig.~\ref{fig:prox}. $\texttt{FedProx}$ does not show a significant improvement in the model's performance compared to \FedAvg. Our method $\yolo$ continues to demonstrate a substantial advantage. We conclude that $\yolo$ outperforms recent methods designed for federated learning, offering superior performance without the need for fine-tuning optimization parameters. $\texttt{FedDyn}$ propose a dynamic regularizer for each device at each round. The results are similar. It should also be mentioned that $\yolo$ can be easily combined with those optimization-based methods.

\subsection{Comparison to \Head} 
\begin{figure*}[t]
\centering
\vspace{-10pt}
\begin{subfigure}{2.1in}
    \centering
	\begin{tikzpicture}
		\node at (0,0) [scale=0.29]{\includegraphics{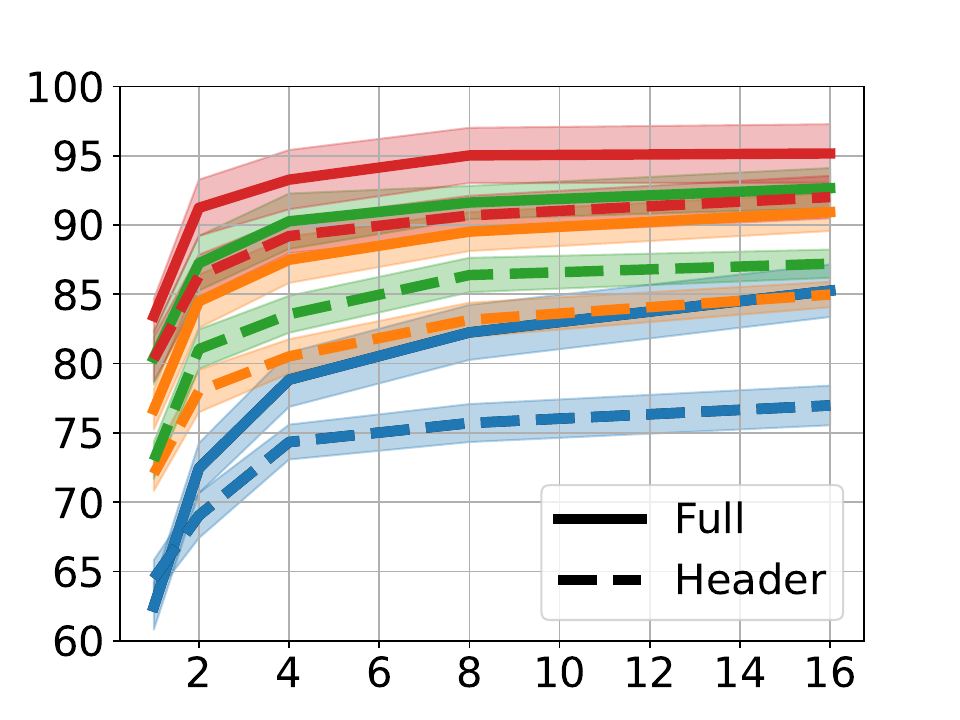}};
		\node at (-2.3 ,0) [scale=0.8,rotate=90] {Accuracy};
  \node at (0,-1.8) [scale=0.8] {\# of samples};
\end{tikzpicture}\vspace{-6pt}\caption{\small{Homogeneous, \FedAvg:\\ \fullup vs. \head}}\label{fig:header_hom_fullhead}
\end{subfigure}
\begin{subfigure}{2in}
    \centering
	\begin{tikzpicture}
		\node at (0,0) [scale=0.29]{\includegraphics{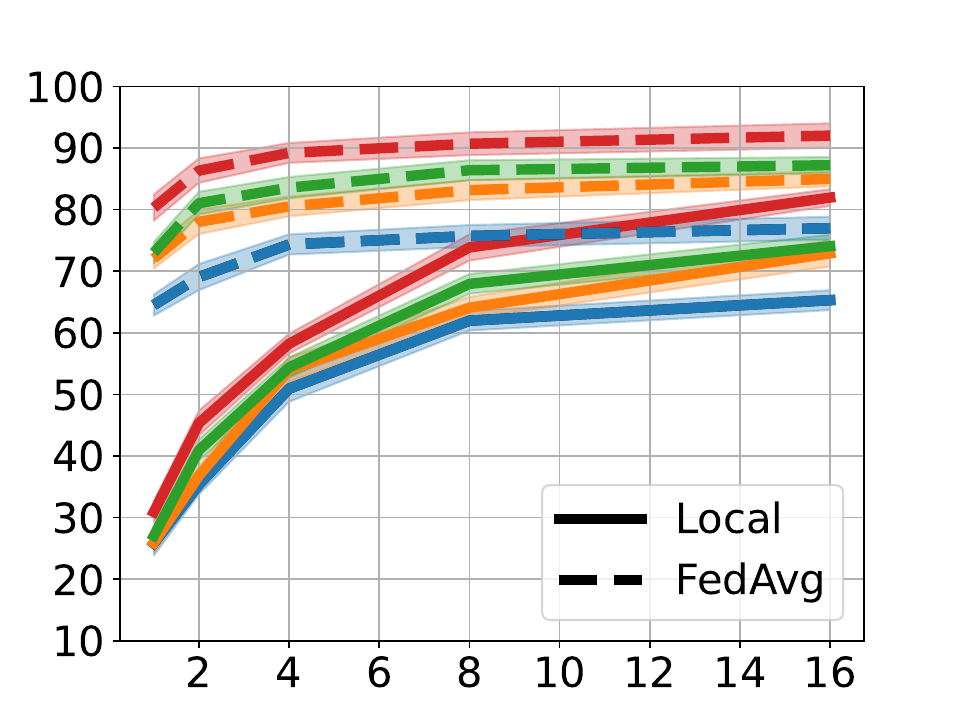}};
  \node at (0,-1.8) [scale=0.8] {\# of samples};
\end{tikzpicture}\vspace{-6pt}\caption{\small{Homogeneous, \head: \\\FedAvg vs. $\texttt{Local-only}$}}\label{fig:header_hom}
\end{subfigure}
\hspace{-22pt}
\begin{subfigure}{0.7in}
    \centering
	\begin{tikzpicture}
\end{tikzpicture}\vspace{60pt}
\end{subfigure}\\
\begin{subfigure}{2.1in}
    \centering
	\begin{tikzpicture}
		\node at (0,0) [scale=0.29]{\includegraphics{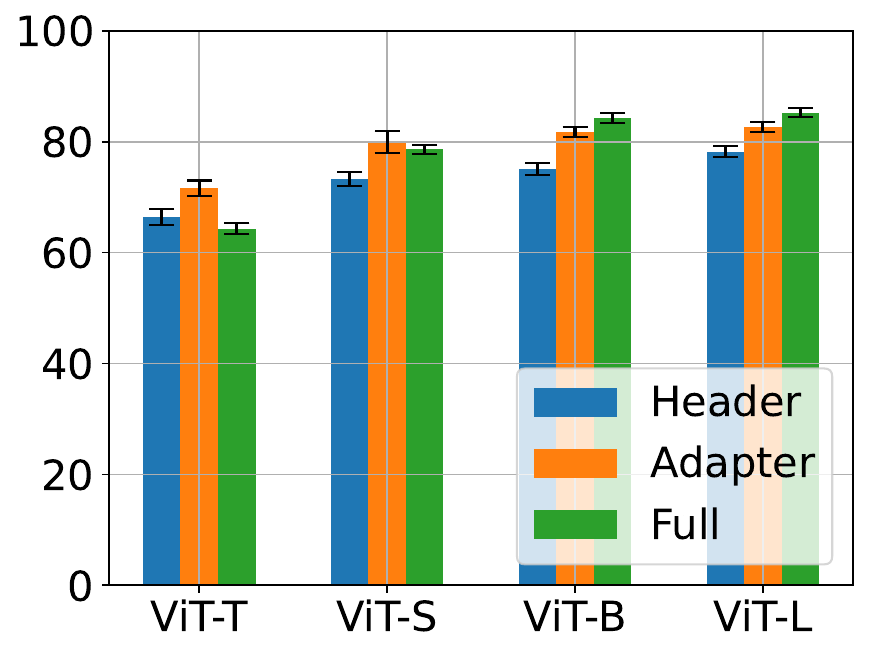}};
		\node at (-2.3 ,0) [scale=0.8,rotate=90] {Accuracy};
\end{tikzpicture}\vspace{-6pt}\caption{\small{Heterogeneous, \FedAvg}}\label{fig:header_heter}
\end{subfigure}
\begin{subfigure}{2in}
    \centering
	\begin{tikzpicture}
		\node at (0,0) [scale=0.29]{\includegraphics{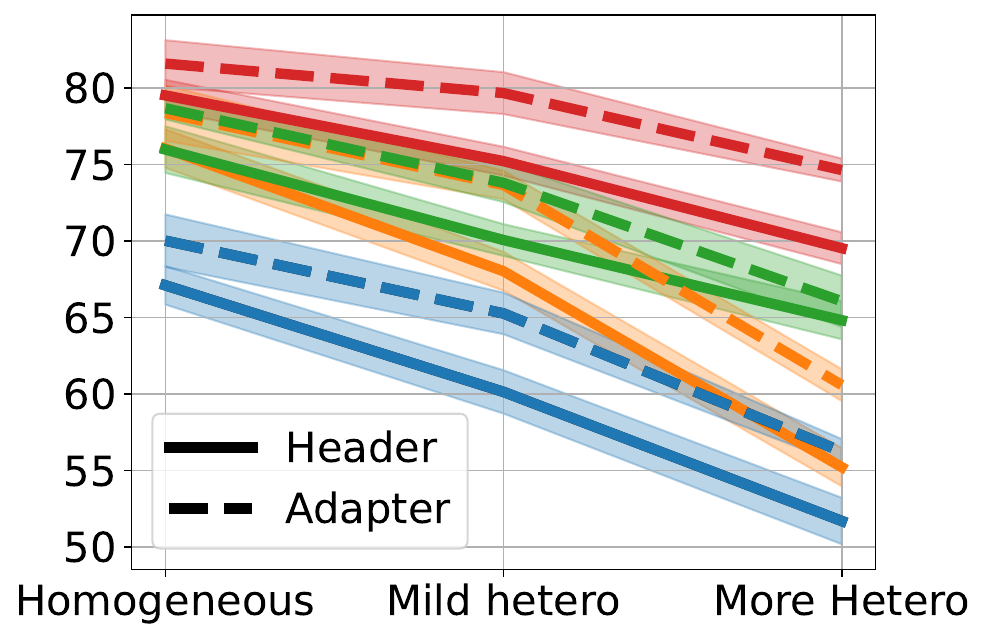}};
\end{tikzpicture}\vspace{-6pt}\caption{\small{Robustness}}\label{fig:header_robustness}
\end{subfigure}
\hspace{-22pt}
\begin{subfigure}{0.7in}
    \centering
	\begin{tikzpicture}
		\node at (0,0) [scale=0.15]{\includegraphics{img/legend.pdf}};
\end{tikzpicture}\vspace{60pt}
\end{subfigure}
\caption{The results of \head. (a,b) Accuracy as a function of the number of training samples per class (CIFAR-100, all clients with
100 classes). Same as the setting in Fig.~\ref{fig:fully_local}(a) Comparing \head (dashed) and \fullup(solid)
training strategies in the federated setting. (b) Comparing the \FedAvg(dashed) with $\local$(solid). (c,d) Experiments are conducted with the mild heterogeneous CIFAR-100 dataset. (c) Model performance of \FedAvg, with heterogeneous data distribution. Same as the setting in Fig.~\ref{fig:heter_cifar100}. \modular consistently outperforms \head in terms of performance. (d) Test accuracy under different levels of data heterogeneity. Same as the setting in Fig.~\ref{fig:heter_cifar100_level}. Comparing the proportion of the same curve's descent from left to right, we observe that \modular (dashed) can achieve performance compared to \head(solid).}
\label{fig:header}
\vspace{-10pt}
\end{figure*}

We also evaluate the performance of the \head method and compare it with \modular in our experiments. The results can be found in Figure~\ref{fig:header}. Among all the settings, the results show that the observations from the \modular experiments also hold true for \head. Furthermore, the results consistently demonstrate that \modular outperforms \head. Figures~\ref{fig:header_hom} and~\ref{fig:header_hom_fullhead} illustrate the model performance with homogeneous data, while Figures~\ref{fig:header_heter} and~\ref{fig:header_robustness} compare the model performance and robustness under the heterogeneous setting. In addition to the previous observation, we find that the \head performs worse than \modular in terms of both performance and robustness.\\
\subsection{Results of Other Modules} \label{app:other_module}
\begin{figure*}[t]
\centering
\begin{subfigure}{2.1in}
    \centering
	\begin{tikzpicture}
		\node at (0,0) [scale=0.29]{\includegraphics{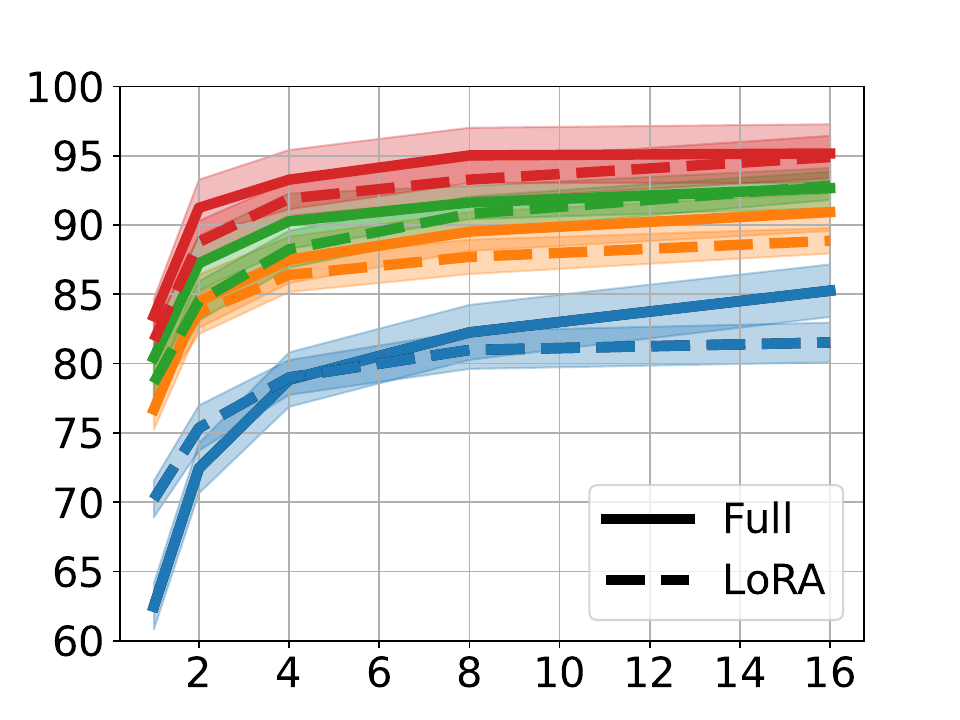}};
		\node at (-2.3 ,0) [scale=0.8,rotate=90] {Accuracy};
\end{tikzpicture}\vspace{-6pt}\caption{\small{Homogeneous, $\FedAvg$:\\ \fullup vs. VPT}}\label{fig:vpt_hom_fullmodular}
\end{subfigure}
\begin{subfigure}{2in}
    \centering
	\begin{tikzpicture}
		\node at (0,0) [scale=0.29]{\includegraphics{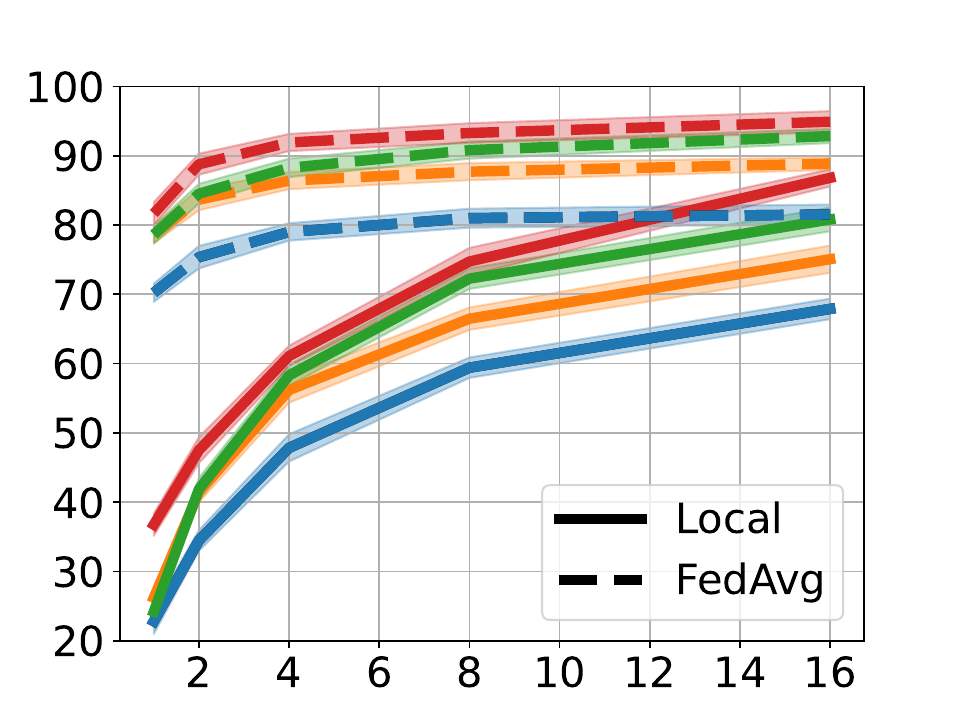}};
\end{tikzpicture}\vspace{-6pt}\caption{\small{Homogeneous, VPT: \\$\FedAvg$ vs. $\texttt{Local-only}$}}\label{fig:vpt_hom_localfed}
\end{subfigure}
\hspace{-22pt}
\begin{subfigure}{0.7in}
    \centering
	\begin{tikzpicture}
		\node at (0,0) [scale=0.15]{\includegraphics{img/legend.pdf}};
\end{tikzpicture}\vspace{60pt}
\end{subfigure}\\
\begin{subfigure}{2.1in}
    \centering
	\begin{tikzpicture}
		\node at (0,0) [scale=0.29]{\includegraphics{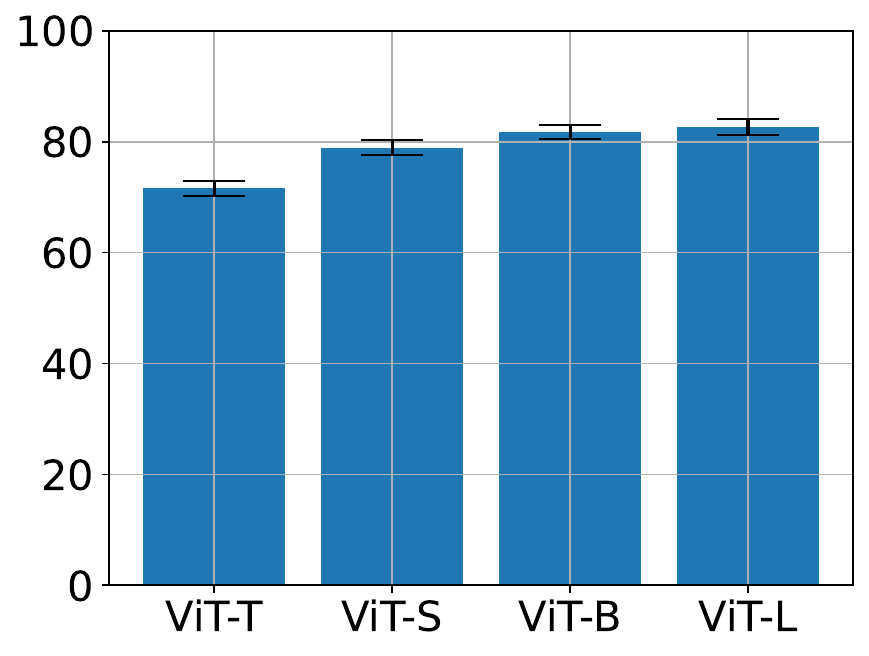}};
		\node at (-2.3 ,0) [scale=0.8,rotate=90] {Accuracy};
\end{tikzpicture}\vspace{-6pt}\caption{\small{VPT: Heterogeneous, $\FedAvg$}}\label{fig:vpt_heter}
\end{subfigure}
\begin{subfigure}{2in}
    \centering
	\begin{tikzpicture}
		\node at (0,0) [scale=0.29]{\includegraphics{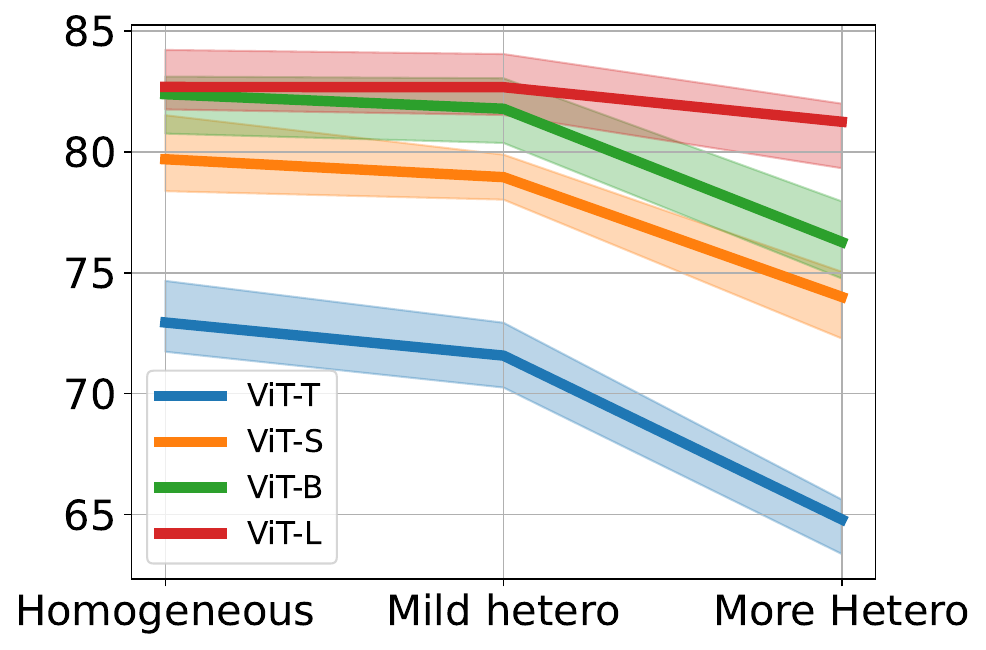}};
\end{tikzpicture}\vspace{-6pt}\caption{\small{VPT: Robustness}}\label{fig:vpt_robustness}
\end{subfigure}
\hspace{-22pt}
\begin{subfigure}{0.7in}
    \centering
	\begin{tikzpicture}
\end{tikzpicture}\vspace{60pt}
\end{subfigure}\\
\begin{subfigure}{2.1in}
    \centering
	\begin{tikzpicture}
		\node at (0,0) [scale=0.29]{\includegraphics{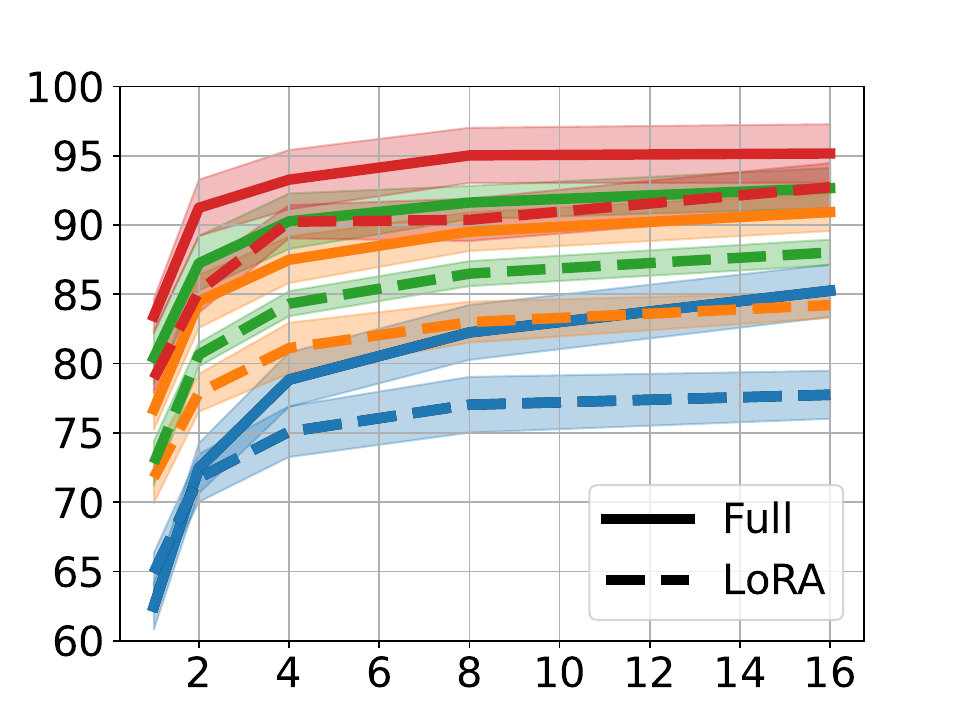}};
		\node at (-2.3 ,0) [scale=0.8,rotate=90] {Accuracy};
\end{tikzpicture}\vspace{-6pt}\caption{\small{Homogeneous, $\FedAvg$:\\ \fullup vs. LoRA}}\label{fig:lora_hom_fullmodular}
\end{subfigure}
\begin{subfigure}{2in}
    \centering
	\begin{tikzpicture}
		\node at (0,0) [scale=0.29]{\includegraphics{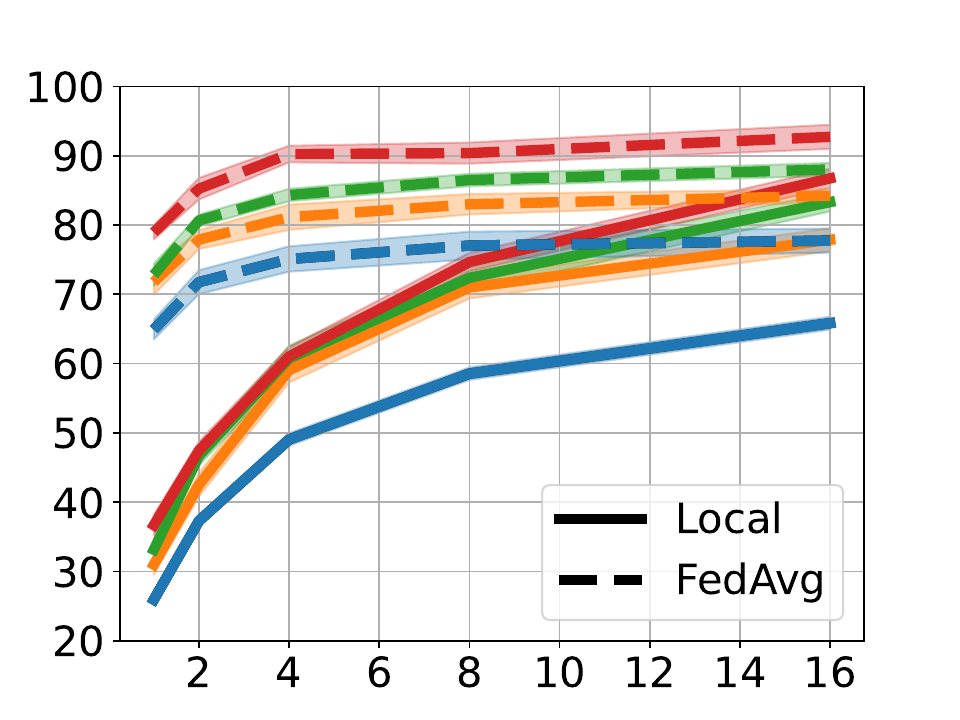}};
\end{tikzpicture}\vspace{-6pt}\caption{\small{Homogeneous, LoRA: \\$\FedAvg$ vs. $\texttt{Local-only}$}}\label{fig:lora_hom_fedlocal}
\end{subfigure}
\hspace{-22pt}
\begin{subfigure}{0.7in}
    \centering
	\begin{tikzpicture}
		\node at (0,0) [scale=0.15]{\includegraphics{img/legend.pdf}};
\end{tikzpicture}\vspace{60pt}
\end{subfigure}\\
\begin{subfigure}{2.1in}
    \centering
	\begin{tikzpicture}
		\node at (0,0) [scale=0.29]{\includegraphics{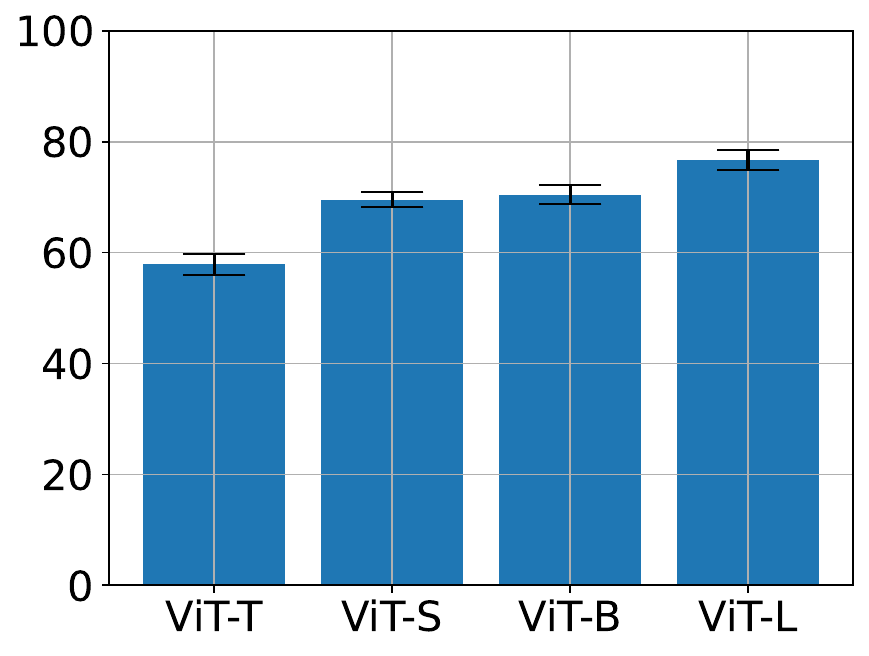}};
		\node at (-2.3 ,0) [scale=0.8,rotate=90] {Accuracy};
\end{tikzpicture}\vspace{-6pt}\caption{\small{LoRA: Heterogeneous, $\FedAvg$}}\label{fig:lora_heter}
\end{subfigure}
\begin{subfigure}{2in}
    \centering
	\begin{tikzpicture}
		\node at (0,0) [scale=0.29]{\includegraphics{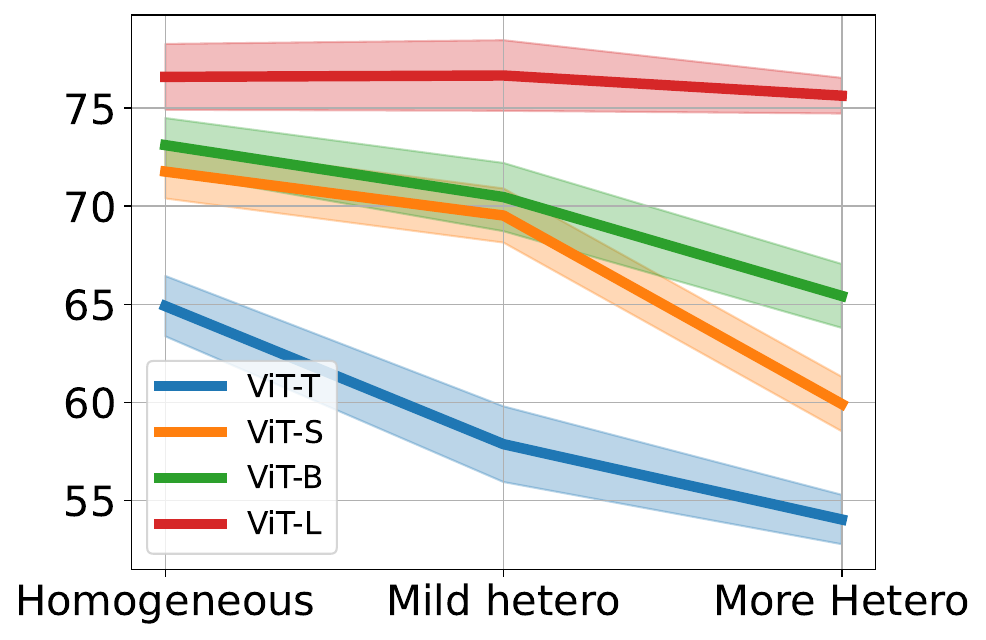}};
\end{tikzpicture}\vspace{-6pt}\caption{\small{LoRA: Robustness}}\label{fig:lora_robustness}
\end{subfigure}
\hspace{-22pt}
\begin{subfigure}{0.7in}
    \centering
	\begin{tikzpicture}
\end{tikzpicture}\vspace{60pt}
\end{subfigure}
\caption{The results of other modules,VPT(a-d) and LoRA(e-h). (a,b,e,f) Accuracy as a function of the number of training samples per class (CIFAR-100, all clients with
100 classes). Same as the setting in Fig.~\ref{fig:fully_local}. (c,d) Experiments are conducted with the mild heterogeneous CIFAR-100 dataset.}
\label{fig:modular}
\end{figure*}

In Figure~\ref{fig:modular}, we present the results of VPT and LoRA. While the type of module does influence the performance, our main findings generalize well across module types and experiments.
\subsection{The Impact of Trainable Parameter Count} 
\begin{figure}[ht]
\centering
\begin{subfigure}{1.7in}
    \centering
	\begin{tikzpicture}
		\node at (0,0) [scale=0.29]{\includegraphics{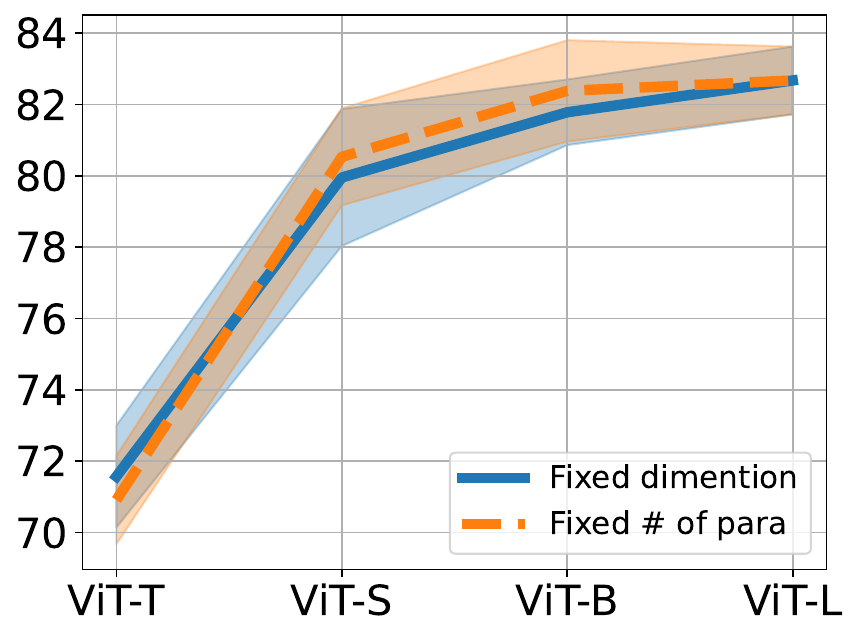}};
  		\node at (-2.3 ,0) [scale=0.8,rotate=90] {Accuracy};
\end{tikzpicture}\vspace{-4pt}\label{fig:ablation_numpara}
\end{subfigure}
\caption{Our experiments demonstrate that increasing the number of parameters for smaller PTFs does not necessarily lead to improved performance. This finding suggests that the advantage of larger PTFs is not due to the larger number of parameters.}
\label{fig:app_para}
\vspace{-5pt}
\end{figure}

To verify that our empirical findings indeed arise from large-scale and modularity rather than other factors, we conduct ablation experiments. Due to the nature of module architectures like LoRA, fixed-size modules across different backbone sizes were not feasible. We use fixed dimension instead of fixed \# of parameters across different scales of PTFs for a fair comparison. We explored the influence of the \# of parameters in the modules using the VPT method. Among the modules, the dimensions of prompts are flexible and can be adjusted accordingly. We vary the dimensions of the VPT while keeping the total number of parameters equal to that of the ViT-L used in our experiments (299,108 parameters). In \cite{lester2021power}, the results indicate that there is a saturation point of prompt size in performance improvement. Beyond that value, further increasing the prompt size does not lead to a significant improvement in performance. Our results(shown in Fig.~\ref{fig:app_para}) also verify the same conclusion. This finding suggests that the advantage of larger PTFs is not attributed to the larger number of parameters.

\subsection{Comparison to Centralized Training}
\begin{figure}[h]
\centering
\begin{subfigure}{1.7in}
    \centering
	\begin{tikzpicture}
		\node at (0,0) [scale=0.29]{\includegraphics{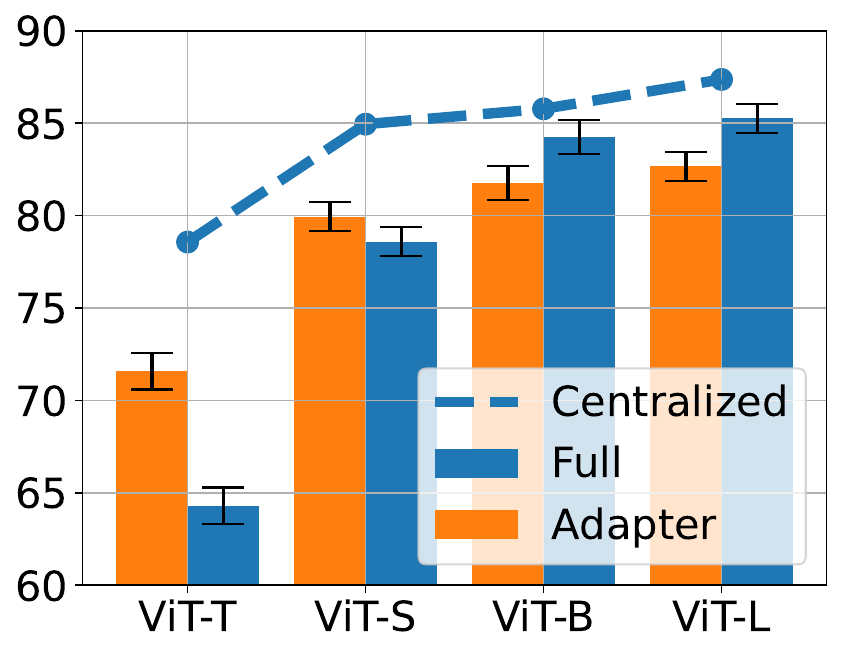}};
		\node at (-2.2 ,0) [scale=0.8,rotate=90] {Accuracy};
\end{tikzpicture}\vspace{-6pt}\caption{\small{CIFAR-100}}\label{fig:app_heter_cifar100}
\end{subfigure}
\begin{subfigure}{1.7in}
    \centering
	\begin{tikzpicture}
		\node at (0,0) [scale=0.29]{\includegraphics{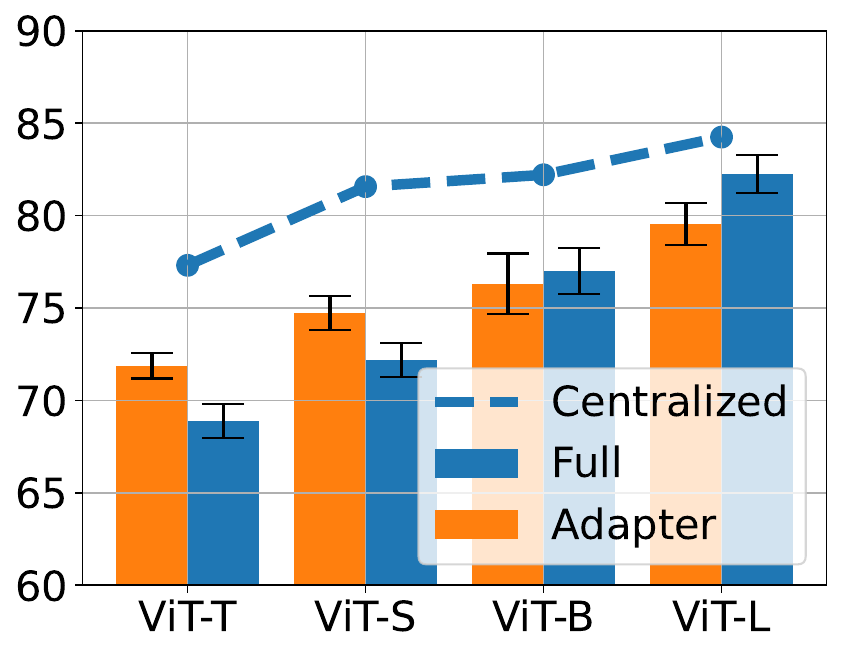}};
\end{tikzpicture}\vspace{-6pt}\caption{\small{CelebA}}\label{fig:app_heter_celeba}
\end{subfigure}
\begin{subfigure}{1.7in}
    \centering
	\begin{tikzpicture}
		\node at (0,0) [scale=0.29]{\includegraphics{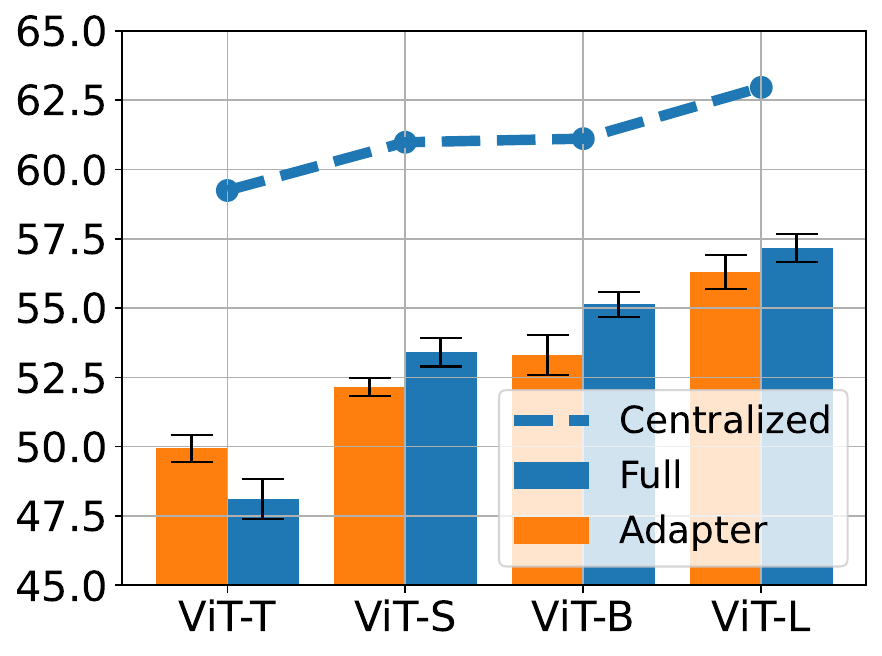}};
\end{tikzpicture}\vspace{-6pt}\caption{\small{FEMNIST}}\label{fig:app_heter_femnist}
\end{subfigure}
\vspace{-5pt}
\caption{\small{Comparison of different models with \FedAvg aggregation and centralized training. The dashed line corresponds to the baseline of full-update centralized training.}}\label{fig:app_heter_acc}
\end{figure}

{
In order to assess the impact of heterogeneity on model performance, we also compare the federated accuracies to the centralized accuracies. The results are shown in Figure~\ref{fig:app_heter_acc}, where we observe that models with larger scale exhibit greater robustness to heterogeneity, consistent with our previous findings.}

\subsection{The Impact of Pretraining}
\begin{figure}[h]
\centering
\begin{subfigure}{1.7in}
    \centering
	\begin{tikzpicture}
		\node at (0,0) [scale=0.29]{\includegraphics{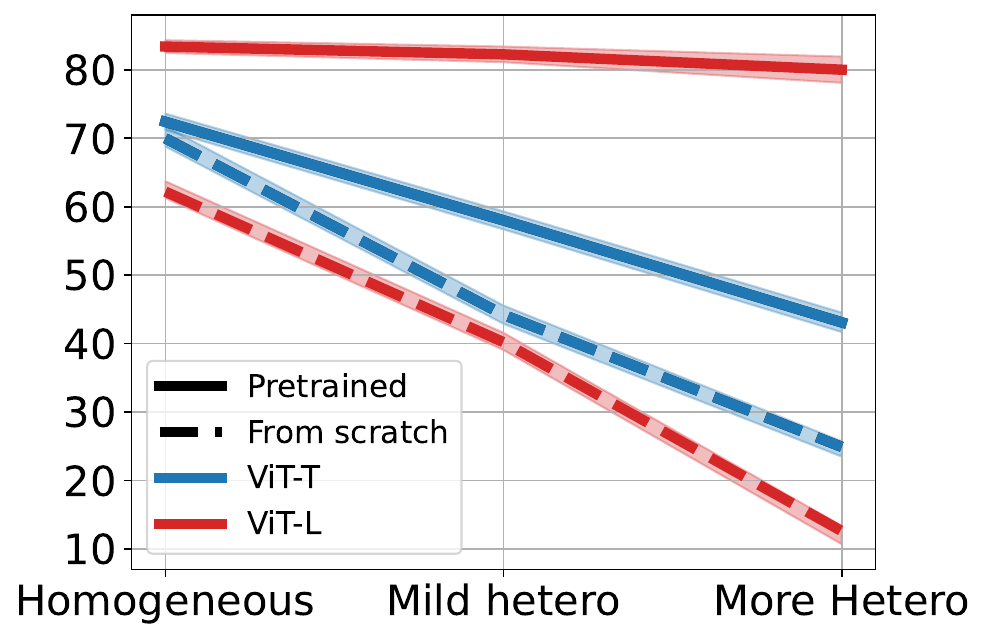}};
  		\node at (-2.3 ,0) [scale=0.8,rotate=90] {Accuracy};
\end{tikzpicture}\vspace{-4pt}
\end{subfigure}
\caption{Our experiments confirm that employing pretrained models in federated learning leads to improved performance compared to models trained from scratch. Furthermore, our findings show that larger-scale models benefit more significantly from pretraining.}
\label{fig:scratch}
\vspace{-5pt}
\end{figure}

{Previous works \cite{weller2022pretrained,qu2022rethinking} experimentally show that using pretrained models could achieve better performance compared to the models trained from scratch for federated learning settings. Our experiments align with these findings and further indicate that larger models tend to benefit more in scenarios where few-shot training is employed. The results are shown in Fig.~\ref{fig:scratch}.  We apply \FedAvg as the training algorithm and test on CIFAR100 with varying levels of heterogeneity.}
\vspace{-0pt}

\section{Experiment details and reproducibility} \label{sec:app_exp_deatil}
We employed a linear learning rate with linear warm-up and cosine decay scheduler for our experiments. In all federated learning methods, we set the local training epoch (E) to 1 (unless otherwise specified) and the total communication rounds to 150. We used the stochastic gradient descent (SGD) optimizer with momentum of 0.9 and no weight decay. The local training batch size was set to 32, and the input image resolution was fixed at 224 × 224 for all methods. For CIFAR experiments, we randomly sampled 5 clients per round, while for FEMNIST and CelebA, we randomly sampled 10\% of clients per round. All experiments were conducted on Tesla V100 or A100 GPU. All the experiments were run for 5 independent runs.
\subsection{Data partition}\label{sec:partition}
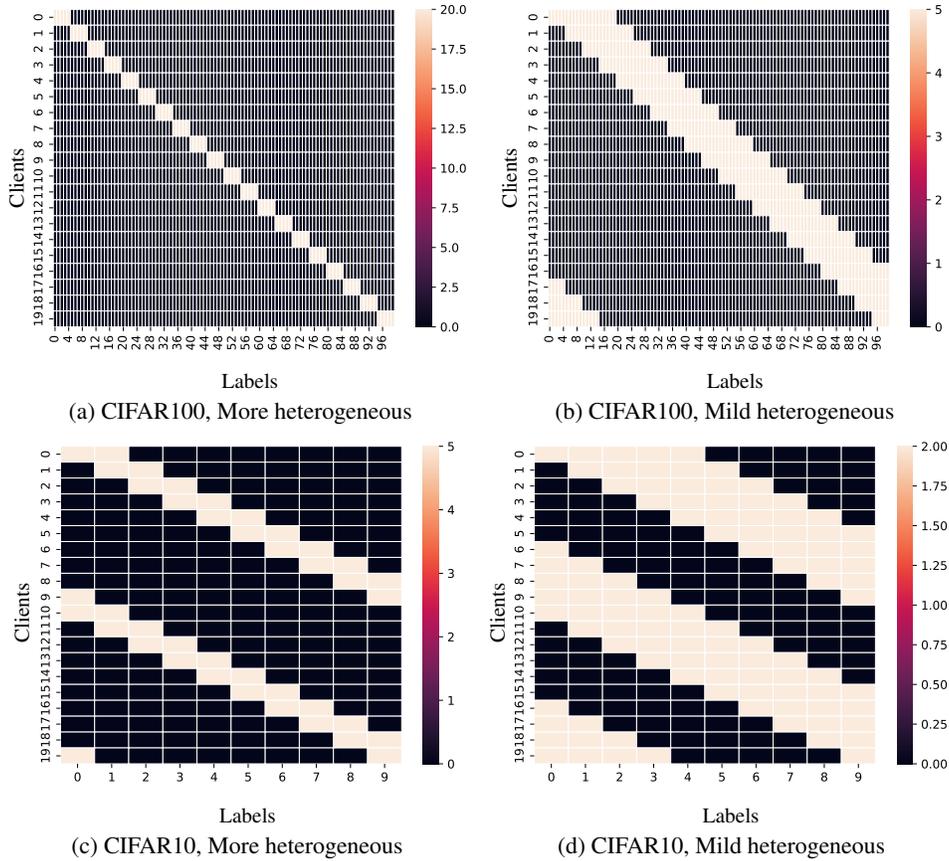
\begin{figure}[h]
\centering
\vspace{-10pt}
\begin{subfigure}{2.5in}
    \centering
	\begin{tikzpicture}
		\node at (0,0) [scale=0.45]{\includegraphics{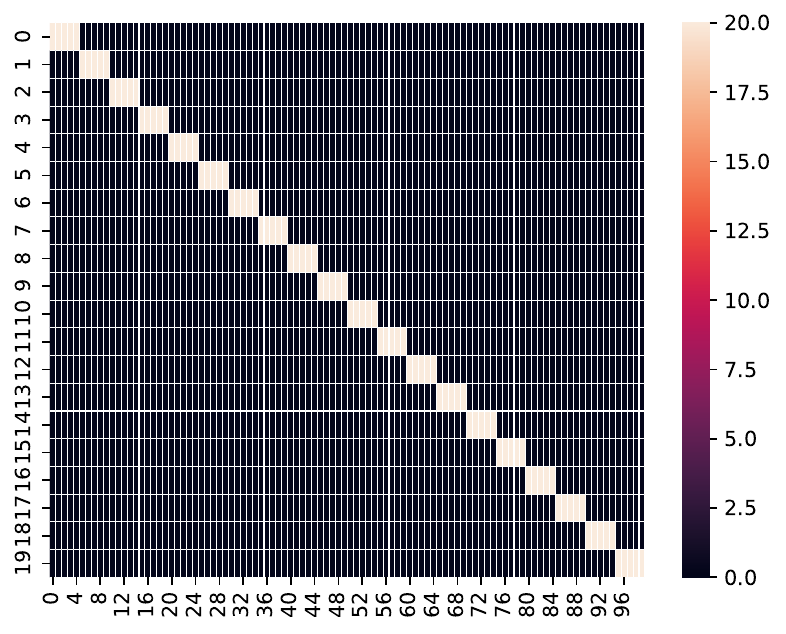}};
  		\node at (-3.1 ,0) [scale=0.8,rotate=90] {Clients};
		\node at (0,-2.7) [scale=0.8] {Labels};
\end{tikzpicture}\vspace{-4pt}\caption{CIFAR100, More heterogeneous}\label{fig:partition_cifar100_5class}
\end{subfigure}
\begin{subfigure}{2.5in}
    \centering
	\begin{tikzpicture}
		\node at (0,0) [scale=0.45]{\includegraphics{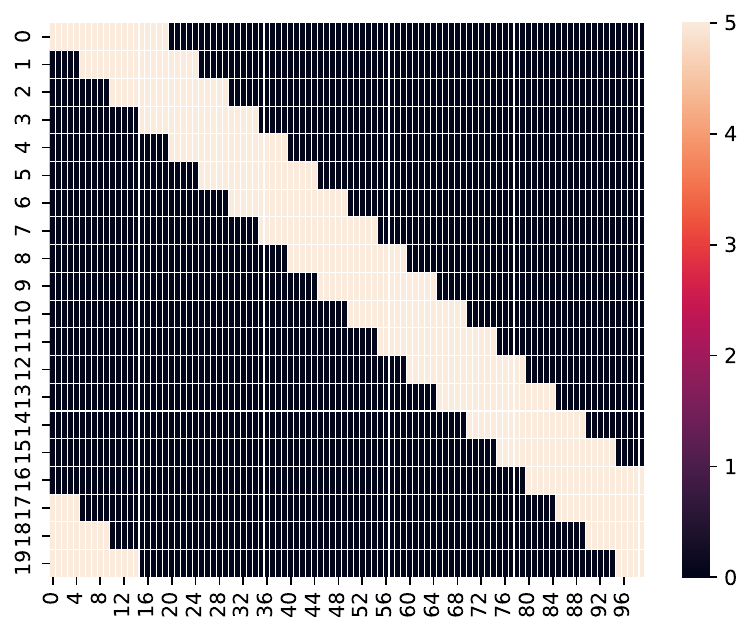}};
		\node at (-3 ,0) [scale=0.8,rotate=90] {Clients};
		\node at (0,-2.7) [scale=0.8] {Labels};
\end{tikzpicture}\vspace{-4pt}\caption{CIFAR100, Mild heterogeneous}\label{fig:partition_cifar100_20class}
\end{subfigure}
\begin{subfigure}{2.5in}
    \centering
	\begin{tikzpicture}
		\node at (0,0) [scale=0.45]{\includegraphics{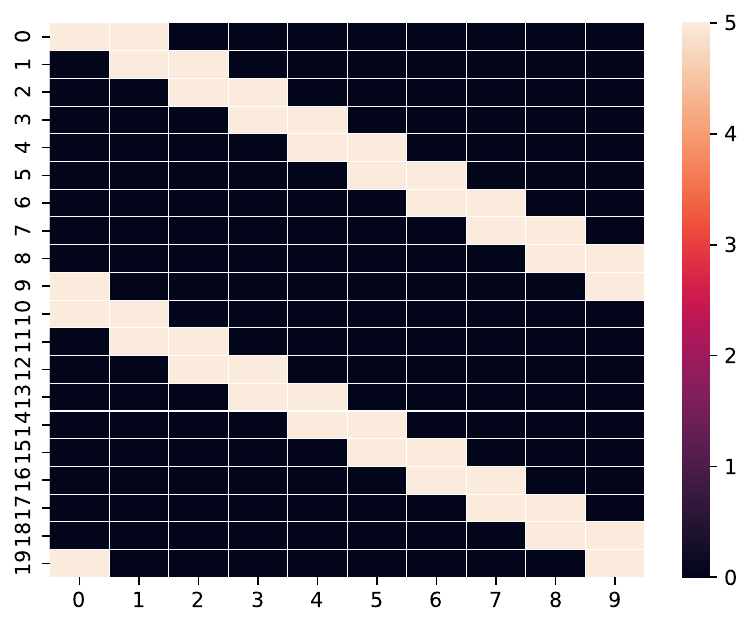}};
  		\node at (-3 ,0) [scale=0.8,rotate=90] {Clients};
		\node at (0,-2.7) [scale=0.8] {Labels};
\end{tikzpicture}\vspace{-4pt}\caption{CIFAR10, More heterogeneous}\label{fig:partition_cifar10_2class}
\end{subfigure}
\begin{subfigure}{2.5in}
    \centering
	\begin{tikzpicture}
		\node at (0,0) [scale=0.45]{\includegraphics{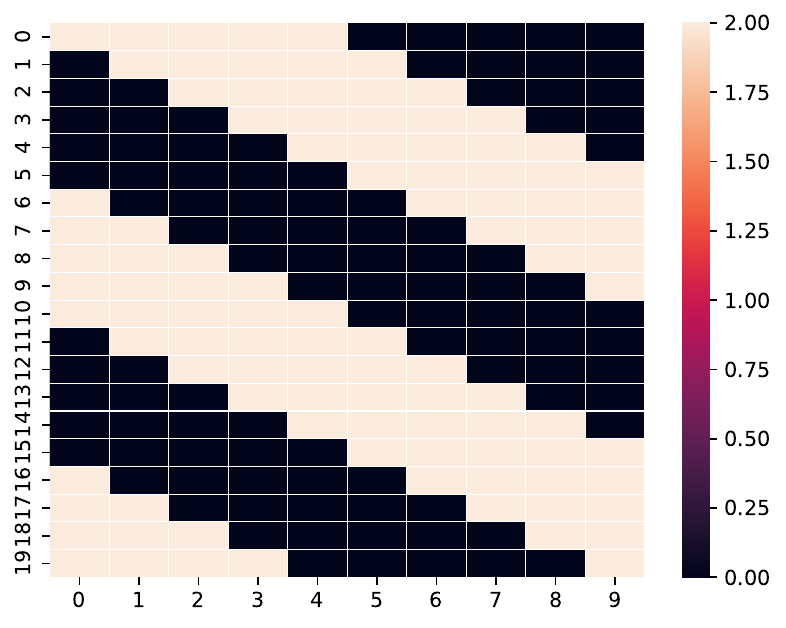}};
		\node at (-3.1 ,0) [scale=0.8,rotate=90] {Clients};
		\node at (0,-2.7) [scale=0.8] {Labels};
\end{tikzpicture}\vspace{-4pt}\caption{CIFAR10, Mild heterogeneous}\label{fig:partition_cifar10_5class}
\end{subfigure}
\vspace{-5pt}
\caption{Data partition for different non-IID level}
\vspace{-4pt}
\label{fig:partition}
\end{figure}

The details of data partition are provided in Fig.~\ref{fig:partition}

\subsection{The setting of modules}\label{sec:module}
The number of trainable parameters for each training strategy was shown in Table.~\ref{tab:num_param_appendix}
\begin{table*}[t]
\centering
\begin{tabular}{lllll}
        & ViT-T & ViT-S & ViT-B & ViT-L \\ \hline
Full model    & 5,543,716 & 21,704,164    &   85,875,556    &   303,404,132 \\
Adapter  &   58,564    &     116,932  &  233,668     &    417,984  \\     
LoRA    &   93,028    &   185,956    &   371,812    &888,932       \\
VPT     &    37,732   &   75,364    &    150,628   &   299,108   \\
Header &  19,300     &  38,500     &  76,900     &   102,500  
\end{tabular}\caption{Number of parameters for different PTF scales and different tuning methods.}\label{tab:num_param_appendix}
\end{table*}
\section{A Representation-based Explanation for the Benefits of Large Models} \label{sec:explanation}
In our discussion, we highlight how the \modular approach and the use of large-scale PTFs contribute to communication efficiency, robustness, and multitask federated learning. These motivate a central question: What is the underlying cause driving these improvements?

\begin{figure}
\centering
\begin{subfigure}{2in}
    \centering
	\begin{tikzpicture}
		\node at (0,0) [scale=0.32]{\includegraphics{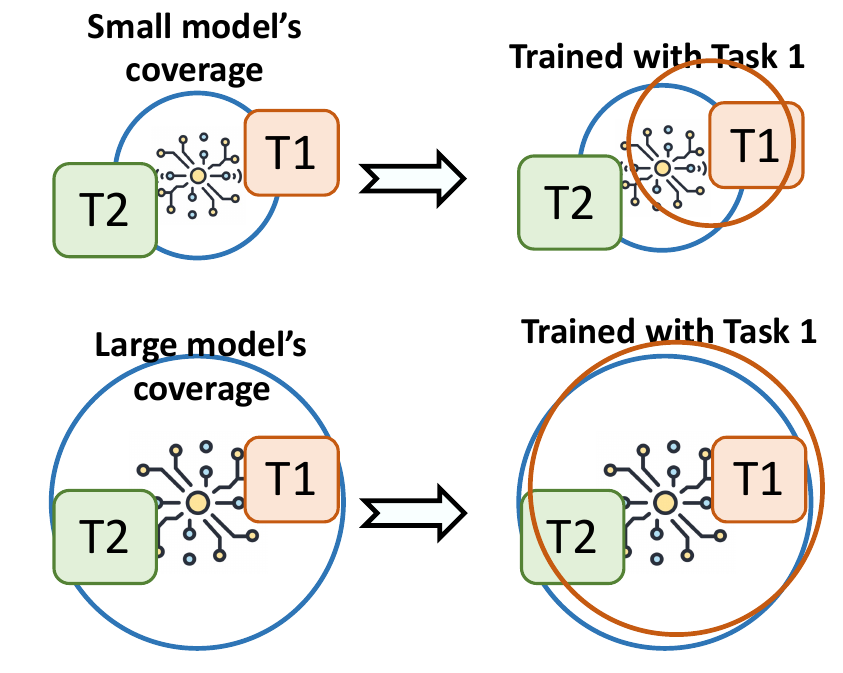}};
\end{tikzpicture}\vspace{-4pt}\caption{}\label{fig:similarity_explain}
\end{subfigure}
\begin{subfigure}{1.7in}
    \centering
	\begin{tikzpicture}
		\node at (0,0) [scale=0.29]{\includegraphics{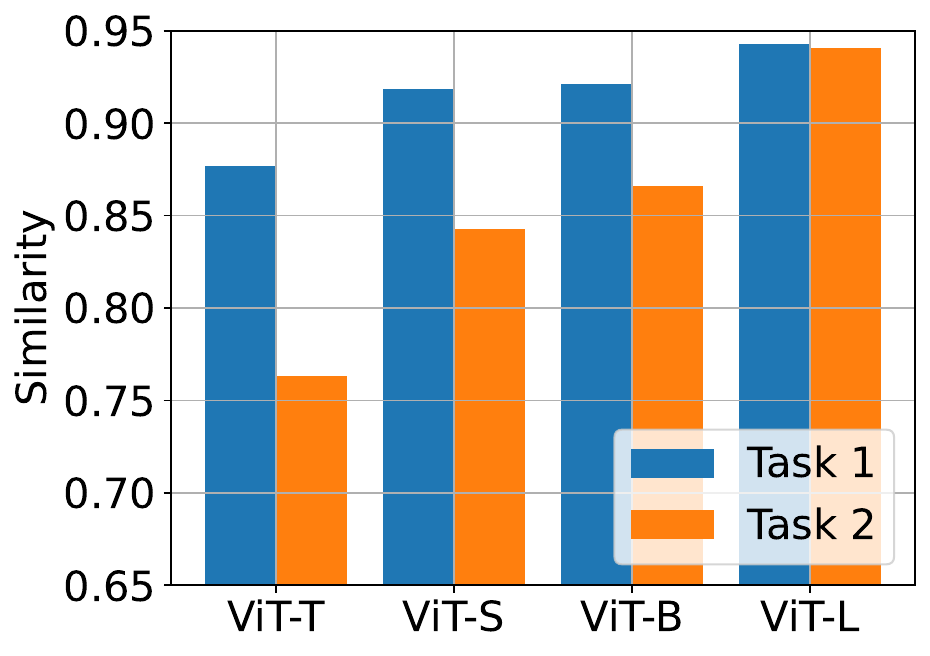}};
\end{tikzpicture}\vspace{-4pt}\caption{}\label{fig:similarity_result_task1}
\end{subfigure}
\caption{(a) The figure illustrates a pre-trained model on ImageNet-21k and its subsequent fine-tuning to a specific task, Task 1. An additional task, Task 2, not involved in training, is also introduced. We employ a bubble analogy to represent the model's captured representation breadth, where larger bubbles symbolize richer features. Notably, larger pre-trained models necessitate minor adjustments to cater to Task 1, thereby implying smaller changes in feature embeddings and minimal interference with Task 2.
(b) Our experiments involved fine-tuning the model on a few-shot CIFAR-100 dataset (400 samples), denoted as Task 1, using the \modular approach. We assessed the feature embedding similarity against both CIFAR-100 (Task 1) and CIFAR-10 (Task 2) test sets. The results exhibit an increasing feature embedding similarity with model size, confirming our hypothesis of smaller feature changes with larger models.}
\label{fig:similarity_result}
\end{figure}

Here, we put forth a plausible explanation: During the process of fine-tuning the model for novel tasks, larger PTFs tend to experience less drastic alterations in their feature embeddings. This concept is illustrated in Figure~\ref{fig:similarity_explain}. The rationale behind this is the comprehensive representation capacity of large PTFs, enabling them to encompass a broad array of features that are effectively generalizable across diverse tasks. Consequently, the federated learning process for these larger PTFs necessitates only minor modifications to the feature embeddings to adapt to distinct clients/tasks, thereby resulting in more effective model updates without significantly altering the original model and forgetting the other clients/tasks.

To support our hypothesis, we conducted experiments which are shown in Fig.~\ref{fig:similarity_result_task1}. Specifically, we compared the changes in feature embeddings during fine-tuning as a function of the PTF scale. For each sample $x$ from a specific task $\mathcal{T}$, we computed the normalized representations $r_x=f(x)$ and $r^{\text{tune}}_x=f^{\text{tune}}(x)$, where $f$ represents the pretrained model and $f^{\text{tune}}$ represents the fine-tuned model.  We use a cosine similarity measure to evaluate the distance between the two representations: 
$$\texttt{Similarity}=\mathbb{E}_{x\in \mathcal{T}}[\texttt{cosine}(r_x, r^{tune}_x)].$$

In Fig.~\ref{fig:similarity_result_task1}, the feature embedding changes for the target task (Task 1) reveals that larger PTFs require smaller modifications to the feature embeddings when adapting to new tasks. This finding also helps explain why larger PTFs are more sample-efficient. Additionally, we extend our analysis to Task 2, which represents another task not included in the model's training. Our experiments show that fine-tuning with the target task does not cause significant changes in the feature embeddings of samples from other tasks, addressing the issue of catastrophic forgetting. Finetuning with large PTFs and modularity minimizes interference with existing knowledge. These observations further support the use of large PTFs and modularity as key factors in achieving improved performance in multi-task federated learning scenarios.


\end{document}